\documentclass[final,journal]{IEEEtran} 

\usepackage{dblfloatfix}      %
\usepackage{mathptmx}         %
\usepackage{helvet}           %
\usepackage[T1]{fontenc} 
\usepackage{courier}          %
\usepackage{type1cm}          %

\usepackage{makeidx}          %
\usepackage{multicol}         %
\usepackage[bottom]{footmisc} %

\usepackage[table]{xcolor}

\usepackage{psfrag}

\usepackage[crop=pdfcrop,mode=batch]{pstool}

\usepackage[sort,compress,space]{cite}

\usepackage{amsmath,empheq,nicefrac,amssymb,bm,mathtools}

\usepackage[ruled,vlined]{algorithm2e}

\usepackage{multirow}
\usepackage{booktabs}
\usepackage{arydshln}
\usepackage{float}

\usepackage[nocomma]{optidef}

\usepackage{dutchcal}

\usepackage[bookmarks=true,breaklinks=true]{hyperref}
\definecolor{mygreen}{rgb}{0,0.6,0}
\definecolor{mygrey}{rgb}{.29,.29,.29}
\definecolor{myblue}{rgb}{.12,.46,.70}
\hypersetup
{colorlinks=true,
	linkcolor=red,
	citecolor=mygreen,
	filecolor=magenta,
	urlcolor=myblue,
	linktoc=page,
	pdfkeywords={constrained systems} {holonomic constraint} {kinodynamic planning}
	{motion planning},
}

\DeclareMathAlphabet{\pazocal}{OMS}{zplm}{m}{n}

\newcommand{\Id}[1]{\ensuremath{\mt{I}_{#1}}}

\newcommand{\R}[1]{\ensuremath{\mathbb{R}^{#1}}}

\newcommand{\vr}[1]{{\mbox{\bm{$#1$}}}}

\newcommand{\mt}[1]{{\mbox{\bm{$#1$}}}}

\newcommand{\manif}[1]{\ensuremath{\pazocal{#1}}}

\newcommand{\Link}[1]{\ensuremath{\pazocal{L}}_{#1}}

\newcommand{\X}{\ensuremath{\manif{X}}} %

\newcommand{\TX}[1]{\ensuremath{\pazocal{T}_{#1}\X}}

\newcommand{\tr}{^{\!\top}}

\newcommand{\q}{\ensuremath{\vr{q}}}
\newcommand{\dq}{\ensuremath{\vr{\dot{q}}}}
\newcommand{\ddq}{\ensuremath{\vr{\ddot{q}}}}

\title{
	\vspace{-15mm}
	\begin{center}
		{\small The final version of this paper was published in IEEE Trans. on Robotics, Volume: 39, Issue: 1, February 2023} \\ [-.7em]
		{\small DOI link: \url{https://doi.org/10.1109/TRO.2022.3193776}} \\ [.5em]
		Direct Collocation Methods for Trajectory \\
		Optimization in Constrained Robotic Systems
	\end{center}
}

\author{Ricard Bordalba, Tobias Schoels, Llu\'{\i}s Ros,
	 Josep M. Porta, and Moritz Diehl%
\thanks{This work has been partially funded by the Spanish Ministry of Science,
	 Innovation, and Universities under projects DPI2017-88282-P, 
	 and PID2020-117509GB-I00/AEI/10.13039/50110001103,
	 and by  the German Federal Ministry for Economic 
   Affairs and Energy (BMWi) via DyConPV (0324166B).} 
\thanks{Ricard Bordalba, Llu\'{\i}s Ros, and Josep M. Porta are with the
Institut de Rob\`otica i Inform\`atica Industrial (CSIC-UPC). 
C. Llorens Artigas 4-6, 08028 Barcelona, Catalonia
(e-mails: rbordalba@iri.upc.edu, ros@iri.upc.edu, and porta@iri.upc.edu).}
\thanks{Tobias Schoels is with the
	Systems Control and Optimization Laboratory, 
	Department of Microsystems Engineering (IMTEK),
	University of Freiburg, Georges-Koehler-Allee 102, 79110 Freiburg, Germany
	(e-mail: tobias.schoels@imtek.de).}
\thanks{Prof. Dr. Moritz Diehl is with the
Systems Control and Optimization Laboratory, 
Department of Microsystems Engineering (IMTEK) and Department of Mathematics, 
University of Freiburg, Georges-Koehler-Allee 102, 79110 Freiburg, Germany 
(e-mail: moritz.diehl@imtek.uni-freiburg.de).}
}

\begin{document}

\maketitle
\thispagestyle{empty}
\pagestyle{empty}

\begin{abstract}
Direct collocation methods are powerful tools to solve trajectory optimization
problems in robotics. While their resulting trajectories tend to be dynamically
accurate, they may also present large kinematic errors in the case of
constrained mechanical systems, i.e., those whose state coordinates are subject
to holonomic or nonholonomic constraints, like loop-closure or rolling-contact
constraints. These constraints confine the robot trajectories to an
implicitly-defined manifold, which complicates the computation of
accurate solutions. Discretization errors inherent to the transcription of the
problem easily make the trajectories drift away from this manifold, which results
in physically inconsistent motions that are difficult to track with a
controller. This paper reviews existing methods to deal with this problem and
proposes new ones to overcome their limitations. Current approaches either
disregard the kinematic constraints (which leads to drift accumulation) or
modify the system dynamics to keep the trajectory close to the manifold (which
adds artificial forces or energy dissipation to the system). The methods we
propose, in contrast, achieve full drift elimination on the discrete trajectory,
or even along the continuous one, without artificial modifications of the system
dynamics. We illustrate and compare the methods using various examples of
different complexity.

\end{abstract}

\section{Introduction} 
\label{sec:intro} 

\IEEEPARstart{T}{rajectory} synthesis is one of the most fundamental problems in
robotics. The goal is to find a sequence of control actions able to move a robot
from a start to a goal state, while avoiding collisions with obstacles and
incurring in minimum costs during the way. The problem is difficult in general.
The control actions should be carefully selected to ensure the kinodynamic
feasibility of the trajectory, which requires to take many aspects into account.
A full robot model should be accounted for---including inertial and friction
effects, motor saturations, joint or velocity limits, and other kinematic or
dynamic constraints of relevance---as well as a rich-enough model of the
environment. Traditionally, the robotics community has approached this problem
with motion planners of various sorts, like those using potential fields,
probabilistic roadmaps, or randomized tree techniques, among others
\cite{Lavalle_06}. Recently,
however, the %
advances in computing power and mathematical programming are also giving rise to
a new family of planners based on trajectory optimization
methods~\cite{Kalakrishnan_2011,Zucker_IJRR2013,Schulman_IJRR2014}. A strong
point of these methods is versatility. Constraints and cost functions of various
forms can all be managed under a same paradigm, allowing to design fast, smooth,
or control-efficient motions in broad classes of systems.

In essence, all trajectory optimization methods solve an instance of the
variational problem of optimal control. Two strategies are mainly followed
\cite{benson2006direct}. Indirect approaches initially derive the Pontryagin
conditions of optimality and then solve the resulting boundary-value problem
numerically. Direct approaches, in contrast, discretize the optimal control
problem at the outset, and then tackle the discrete problem with nonlinear
programming (NLP) methods. While indirect approaches tend to be more accurate on
optimizing the cost function, they also require good guesses of the solution,
which are difficult to provide in general. Direct approaches, in contrast, may
yield slightly suboptimal trajectories, but show larger regions of convergence,
which makes them preferable to solve problems in robotics. In these approaches,
the dynamic constraints can be discretized using shooting methods, which often
use explicit integrators to estimate the evolution of the system, or collocation
methods, which avoid costly integration rules via spline approximations.
Collocation methods are relatively fast and effectively solve a wide variety of
problems, which justifies the growing interest they
arouse~\cite{Kelly_SIAM2017,Posa_ICRA2016,Pardo_RSS17,
	Posa_IJRR2014,Patel-RAL2019}, and why they constitute the main focus of this
paper.

When computing a solution trajectory, a main concern is the consistency of the
motor actions with the trajectory states, so the actions can closely reproduce
such states when executed in the real robot. This calls for the use of realistic
robot models, but also for an accurate satisfaction of the kinematic and dynamic
constraints of such models along the entire trajectory. While direct collocation
methods are good at achieving dynamic accuracy, little work has been devoted to
also ensure their kinematic accuracy on constrained mechanical systems; i.e.,
systems with dependent state coordinates, like those involving closed kinematic
chains, rolling contacts, or non-minimal representations of spatial
orientations. In such systems, the state vector must satisfy a set of holonomic
and nonholonomic constraints, which in general cannot be solved in closed form.
This confines the robot trajectories to a nonlinear manifold implicitly defined
in a larger ambient space, so any drift from such manifold caused by
discretization errors will lead to unrealistic trajectories that are difficult
to stabilise with a controller~\cite{Posa_IJRR2014,Posa_ICRA2016}.

\begin{figure}[t!]
	\begin{center}
		\pstool[width=\linewidth]{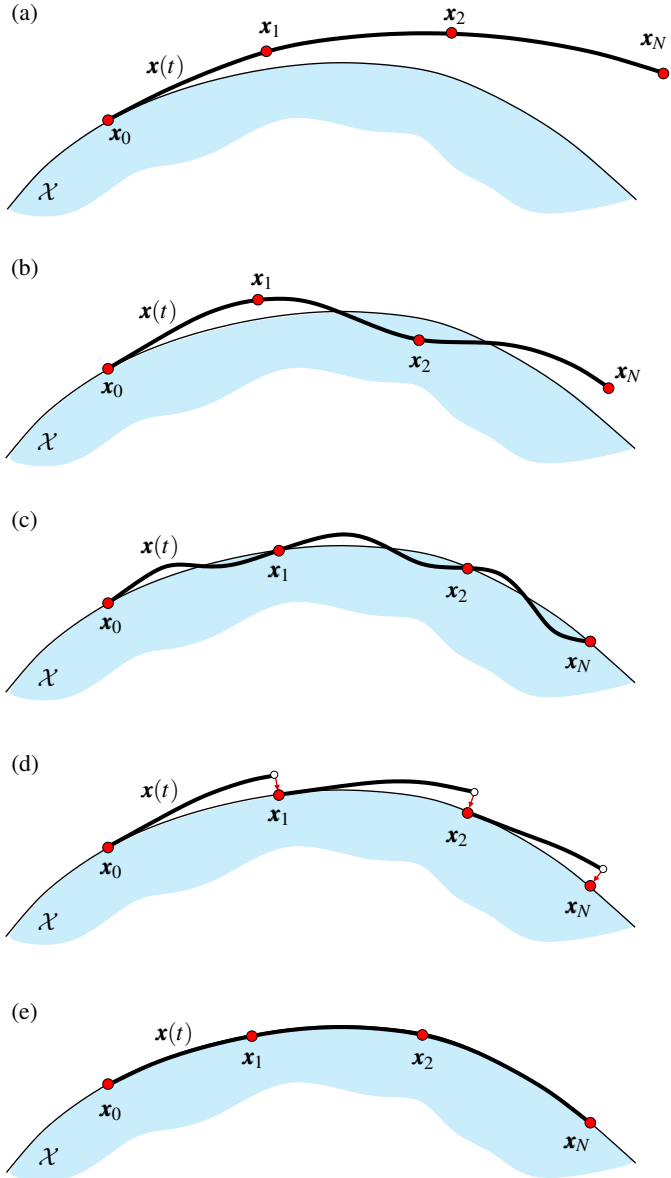} { \psfrag{a}[l]{\small (a)}
	\psfrag{b}[l]{\small (b)} \psfrag{c}[l]{\small (c)} \psfrag{d}[l]{\small (d)}
	\psfrag{e}[l]{\small (e)} \psfrag{X}[l]{\small $\X$} \psfrag{xg}[l]{\small
		$\vr{x}_{g}$} \psfrag{x0}[l]{\small $\vr{x}_{0}$} \psfrag{x1}[l]{\small
		$\vr{x}_{1}$} \psfrag{x2}[l]{\small $\vr{x}_{2}$} \psfrag{xn}[l]{\small
		$\vr{x}_{N}$} \psfrag{xng}[l]{\small $\vr{x}_{N} = \vr{x}_{g}$}
	\psfrag{xt}[l]{\small $\vr{x}(t)$} } 
	\caption{Qualitative form of the trajectories $\vr{x}(t)$ obtained by existing
	methods (a, b, and c) and those proposed in this paper (d and e). From top to
	bottom: output of the ``basic'', ``Baumgarte'', ``PKT'', ``projection'', and
	``local coordinates'' methods. The blue surface $\X$ is the state space
	manifold on which $\vr{x}(t)$ should evolve. The red dots indicate the knot
	states $\vr{x}_0, \ldots, \vr{x}_N$ used to discretize $\vr{x}(t)$. In the
	basic method, the knot states, and so $\vr{x}(t)$, may increasingly drift away
	from $\X$. The knot states are kept near $\X$ in the Baumgarte method (b), and
	exactly on $\X$ in the PKT method (c), but at the cost of modifying the system
	dynamics. The methods we propose also eliminate drift at the knot states (d),
	or even along the entire trajectory (e), without modifying the system
	dynamics.\label{fig:dae_methods} 
	}
	\end{center}
\end{figure}

In the literature, three main strategies have been given to mitigate the problem
of drift. In a first strategy, which we call the basic method, the kinematic
constraints are simply not enforced, but fine discretizations or high-order
integrators can be used to keep trajectory drift to a
minimum~\cite{Felis_Humanoids2015,
	Xi_IJRR2016,Xi_IROS2014,Bonalli_RSS2019,Pardo_RSS17,Patel-RAL2019}. This
approach is easy to implement, but it increases the computational cost of
solving the problem, and much drift can accumulate despite the precautions
\cite{Hairer_NM2001,Gros_CDC2015}. In a second strategy, Baumgarte stabilization
is used to modify the dynamic vector field of the system and make it convergent
to the manifold \cite{Baumgarte_CMAME1972,rabier2002theoretical}. The method is
also easy to implement, but it has three shortcomings: it adds artificial
compliance and energy dissipation to the system, its parameters may be difficult
to tune, and trajectory instabilities may arise as a
result~\cite{rabier2002theoretical,Blajer_CMAME2011}. A third strategy, which we
call the PKT method, adapts the classical Hermite-Simpson collocation scheme to
keep the trajectory knot points on the manifold \cite{Posa_ICRA2016}. This
approach is a notable improvement over the Baumgarte method, but it also applies
artificial modifications of the system dynamics in half of the collocation
points, so the obtained trajectories may suffer from dynamic inconsistencies.
Another drawback is that it can only use cubic polynomials, so it cannot be
applied in hp-adaptive meshing schemes that improve the trajectory by increasing
the polynomial degree, in addition to the mesh resolution
\cite{darby2011hp,Betts_SIAM2010}.

The goal of this paper is to review the previous methods and to propose
new ones to overcome their limitations. By using projections and local charts of
the manifold, we present two methods that keep the trajectory knot points
exactly on the manifold without modifying the system dynamics
(Fig.~\ref{fig:dae_methods}). The two methods, which we refer to as the
``projection'' and ``local coordinates'' methods, can compute approximation
polynomials of arbitrary degree. While the former is easier to implement and
usually faster, the latter achieves full drift elimination even for the
continuous-time trajectory, which is beneficial when high-quality solutions
are needed.

Both methods leverage techniques to solve differential algebraic equations
(DAEs) that can be expressed as ODEs on
manifolds~\cite{Hairer_NM2001,Hairer_SPRINGER2006}. The projection method is
based on the work by Hairer \cite{hairer1996solving}, who used orthogonal
projections to cancel the drift from the algebraic manifold of a DAE. The local
coordinates method, in turn, has its roots in Potra and
Rheinboldt~\cite{potra1991numerical}, and Yen et al.~\cite{Potra_JSM1991}, who
introduced the tangent space parameterization as a means to obtain trajectories
lying continuously on such manifold. While the work
in~\cite{Hairer_NM2001,Hairer_SPRINGER2006,
	hairer1996solving,potra1991numerical,Potra_JSM1991} was mainly developed for
dynamic system simulation, our goal is to show how it can be adapted to
trajectory optimization in robotics.

The rest of the paper is structured as follows. Section~\ref{sec:formulation}
formally describes the class of systems we consider and formulates the
continuous optimal control problem to be solved. Section~\ref{sec:background}
recalls background techniques to transcribe this problem into a nonlinear
programming problem. Using these techniques,
Section~\ref{sec:conventional_schemes} reviews the basic, Baumgarte, and PKT
methods and discussses their strengths and weaknesses.
Section~\ref{sec:new_schemes} explains the methods we propose, which
overcome such weaknesses. Important points to be considered in their
implementation are then given in Section \ref{sec:implementation_details}, and
the performance of all methods is compared in Section~\ref{sec:examples} by
means of examples. Section~\ref{sec:conclusion} finally provides the conclusions
of the paper and highlights points deserving further attention.

\section{Problem formulation and assumptions}  
\label{sec:formulation}

Let us describe the robot state by means of a tuple $\vr{x} = (\q, \dq)$, where
$\q$ is a vector of~$n_q$ generalized coordinates encoding the positions and
orientations of all links at a given instant of time, and $\dq =
\frac{d}{dt}\q$. We restrict our attention to constrained robotic systems, i.e.,
those in which $\q$ and $\dq$ are subject to equations of the form
\begin{subequations}
	\label{eq:kinematic_eqs}
	\begin{empheq}[left=\empheqlbrace]{align}
		& \mt{\Phi}(\vr{q}) = \vr{0}, \label{eq:phi} \\ 
		& \mt{B}(\vr{q}) \; \dq = \vr{0}, \label{eq:Bqdot}   
	\end{empheq} 
\end{subequations} 
\\where $\vr{\Phi}(\vr{q}):\R{n_q}\rightarrow\R{n_p}$ and $\mt{B}(\vr{q})
\;\dq:\R{2n_q}\rightarrow\R{n_v}$ are differentiable maps defining configuration
and velocity constraints. Here, Eq. \eqref{eq:phi} encompasses the holonomic
constraints of the system (like joint-assembly or loop-closure constraints), and
Eq.~\eqref{eq:Bqdot} includes the time derivative of Eq. \eqref{eq:phi} and all
nonholonomic constraints intervening (like those arising from rolling contacts).
Thus, $n_v = n_p + n_r$, where $n_r$ is the number of non-holonomic constraints
of the robot.

For ease of explanation, let us write the system in~\eqref{eq:kinematic_eqs} as
\begin{equation} \label{eq:F}
\vr{F}(\vr{x})=\vr{0},
\end{equation}
so $\vr{F}:\R{n_x}\rightarrow\R{n_e}$, where $n_x = 2\;n_q$, and $n_e = n_p
+ n_v$. The state space of the robot is then the set
\begin{equation} 
\X = \{ \vr{x} : \vr{F}(\vr{x})=\vr{0} \}.
\label{eq:manifold}
\end{equation}
In the rest of the paper, we assume that $\vr{F}_{\small \vr{x}}=\partial
\vr{F}/\partial \vr{x}$ is full rank for all $\vr{x}\in \X$, so $\X$ is a smooth
manifold of dimension $d_{\X} = n_x - n_e$. This assumption is not too
restrictive, as geometric singularities can often be removed by judicious
mechanical design~\cite{Bohigas_SPRINGER2017}, or using singularity-avoidance
constraints~\cite{Bordalba_ARK2018,Bohigas_TRO2013,bohigas2016planning}. The
assumption is needed to ensure that the tangent space of $\X$ at $\vr{x}$,
\begin{equation}
\TX{\vr{x}} = \{ \vr{\dot{x}} \in \R{n_{x}} : 
\vr{F}_{\small \vr{x}} \; \vr{\dot{x}} = \vr{0} \},
\label{eq:tangent_space}
\end{equation}
is well-defined and $d_{\X}$-dimensional for any $\vr{x} \in \X$, a property we
shall exploit in the sequel.

We also model the dynamics of the robot using Lagrange's equation with multipliers
\begin{equation} \label{eq:dynamics}
\mt{M}(\vr{q})\:\ddq+\vr{B}(\vr{q})\tr\vr{\lambda}=\vr{\tau}(\vr{u},\vr{q},\dq),
\end{equation}
where $\mt{M}(\vr{q})$ is the positive-definite mass matrix of the system,
$\vr{\lambda}\in \R{n_v}$ is a vector of Lagrange multipliers, $\vr{u}\in
\R{n_u}$ collects the motor forces and torques, and
$\vr{\tau}(\vr{u},\vr{q},\dq) \in \R{n_q}$ accounts
for the generalized Coriolis, gravity, friction, and actuation forces.

Since Eq.~\eqref{eq:dynamics} is a system of $n_q$ equations in $n_q+n_v$
unknowns ($\ddq$ and $\vr{\lambda}$), we need $n_v$ additional constraints in
order to solve it. These can be obtained by taking the time derivative of
Eq.~\eqref{eq:Bqdot}, which yields the acceleration constraint
\begin{equation} \label{eq:dotdotphi}
\vr{B}(\vr{q})\; \ddq = \vr{\xi}(\vr{q},\dq),
\end{equation}
where $\vr{\xi}(\vr{q},\dq) = -\vr{\dot{B}}(\vr{q}) \;\dq$. Eqs.~\eqref{eq:dynamics} 
and~\eqref{eq:dotdotphi} can then be combined into
\begin{equation}
\label{eq:dyn3}
\begin{bmatrix}
\mt{M}(\vr{q}) & \vr{B}(\vr{q})\tr \\ \vr{B}(\vr{q}) & \vr{0}
\end{bmatrix}\begin{bmatrix}
\ddq \\ \vr{\lambda}
\end{bmatrix}=\begin{bmatrix}
\vr{\tau}(\vr{u},\vr{q},\dq) \\ \vr{\xi}(\vr{q},\dq)
\end{bmatrix},
\end{equation}
and, since $\vr{F}_{\small \vr{x}}$ is full rank, so is $\vr{B}(\vr{q})$, and we
can write
\begin{equation}
\label{eq:dyn4}
\ddq=\begin{bmatrix}
\Id{n_q} & \vr{0}
\end{bmatrix}\begin{bmatrix}
\mt{M}(\vr{q}) & \vr{B}(\vr{q})\tr \\ \vr{B}(\vr{q}) & \mt{0}
\end{bmatrix}^{-1}\begin{bmatrix}
\vr{\tau}(\vr{u},\vr{q},\dq) \\ \vr{\xi}(\vr{q},\dq)
\end{bmatrix}.
\end{equation}

By writing the state as $\vr{x}=(\vr{q},\vr{v})$ and adding the constraint $\dq = \vr{v}$, Eq.~\eqref{eq:dyn4} can be expressed in the common  form
\begin{equation}
\label{eq:explicitdynamics} 
\vr{\dot{x}}(t)=\vr{f}(\vr{x}(t),\vr{u}(t)),
\end{equation}
where $\vr{x}(t)$ and $\vr{u}(t)$ are the state and action trajectories. 
These trajectories may be further constrained by a path constraint
\begin{equation}
	\label{eq:p_cons}
	 \vr{h}(\vr{x}(t),\vr{u}(t)) \geq \vr{0}
\end{equation}
that models, for example, joint or force limits, or collision constraints, and by a boundary constraint 
\begin{equation}
	\label{eq:b_cons}
	 \vr{b}(\vr{x}(0),\vr{x}(t_f)) = \vr{0}
\end{equation}
that restricts the values taken by $\vr{x}(t)$ at the initial and final times
$t=0$ and $t=t_f$.

With these definitions, the problem we confront can be posed as follows. Given a
running cost function $L(\vr{x}(t),\vr{u}(t))$, find trajectories
$\vr{x}(t)$ and $\vr{u}(t)$, and a final time $t_f$, that
\begin{mini!}[2] 
	{\vr{x}(\cdot),\vr{u}(\cdot),t_f}
	{\int_{0}^{t_f}L(\vr{x}(t),\vr{u}(t))\;\textrm{d}t}
	{\label{eq:OCP}}
	{}
	\label{eq:OCP_cost} 
	\addConstraint
		{\vr{h}(\vr{x}(t),\vr{u}(t))}{\ge \vr{0},\label{eq:OCP_path}}{t\in [0,t_f],}
	\addConstraint
		{\vr{\dot{x}}(t)}{=\vr{f}(\vr{x}(t),\vr{u}(t)),\quad\label{eq:OCP_dynamics}}{t\in[0,t_f].}
	\addConstraint
		{\vr{b}(\vr{x}(0),\vr{x}(t_f))}{=\vr{0},\label{eq:OCP_boundary}}{}
	\addConstraint
		{t_f\ge}{0.\label{eq:OCP_time}}{}		
\end{mini!}
Note that Eq.~\eqref{eq:F} is not added to this formulation because it is
already accounted for implicitly by Eq.~\eqref{eq:OCP_dynamics}. However, our
goal will be to obtain solutions of this problem that satisfy both \eqref{eq:F}
and~\eqref{eq:OCP_dynamics} as accurately as possible.

\section{Transcription techniques}
\label{sec:background}

\subsection{Problem discretization}

In this paper we solve Problem \eqref{eq:OCP} by transcribing it into an NLP
problem, which entails approximating all functionals in
\eqref{eq:OCP_cost}-\eqref{eq:OCP_dynamics} by functions of discrete states and
actions. To this end, we discretize the time horizon $[0,t_f]$ into $N$
intervals defined by the time instants
\begin{equation} \nonumber
t_0, \ldots, t_k, \dots, t_N,
\end{equation}
where $t_0 = 0$ and $t_N = t_f$, and represent $\vr{x}(t)$
and $\vr{u}(t)$ by the $N+1$ states
\begin{equation} \nonumber
	\vr{x}_0, \ldots, \vr{x}_k, \dots, \vr{x}_N
\end{equation}
and actions
\begin{equation} \nonumber
	\vr{u}_0, \ldots, \vr{u}_k, \dots, \vr{u}_N
\end{equation} 
at those instants. The values $t_0, \ldots, t_N$ are known as the knot points,
and for simplicity we assume them to be uniformly spaced, so $\Delta t =
t_{k+1}-t_k$ takes the same value for \mbox{$k=0,\ldots,N-1$}. If $\Delta t$ is
constant, the time horizon $[0,t_f]$ is fixed, but a variable time horizon
can also be allowed by treating $\Delta t$ as a decision variable of the
problem \cite{Betts_SIAM2010}.

The transcriptions of Eqs. \eqref{eq:OCP_cost} and \eqref{eq:OCP_path} are
relatively straightforward and less relevant for our purposes. They can be done,
for example, by approximating the integral in Eq.~\eqref{eq:OCP_cost} using some
quadrature rule, and by setting $\vr{h}(\vr{x}_k, \vr{u}_k)\geq\vr{0}$ for
$k=0,\ldots,N$. The transcription of \eqref{eq:OCP_dynamics}, in contrast, is
substantially more involved, and will be the main subject of the rest of this
paper. We next recall background techniques to carry it out, which will be
needed hereafter.

\subsection{Transcription of differential constraints}  
\label{subsec:trans_diff_constraint}

To approximate Eq. \eqref{eq:OCP_dynamics}, the first step is to define
$\vr{u}(t)$ in terms of $\vr{u}_0,\dots,\vr{u}_N$. While many choices are
possible here, we use a first-order hold filter due to its good balance between
simplicity and accuracy of representation. For all $t\in
\left[t_k,t_{k+1}\right]$ it will thus be
\begin{equation} \label{eq:actions}
\vr{u}(t) = \vr{u}_k + \frac{t-t_k}{\Delta t} \cdot (\vr{u}_{k+1}-\vr{u}_k).
\end{equation}

The second step is to determine the state $\vr{x}_{k+1}$ that would be reached
from $\vr{x}_k$ under the actions in Eq.~\eqref{eq:actions}. This can be done in many
ways, but in this paper we opt for the Gauss-Legendre
collocation scheme because it has the lowest integration error for a fixed
number of dynamics function evaluations \cite{Hairer_SPRINGER2006}. Our methods,
however, should be easy to adapt to other collocation
schemes~\cite{Kelly_SIAM2017,Betts_SIAM2010}.

\begin{figure}[t!]
	\begin{center}
		\pstool[width=\linewidth]{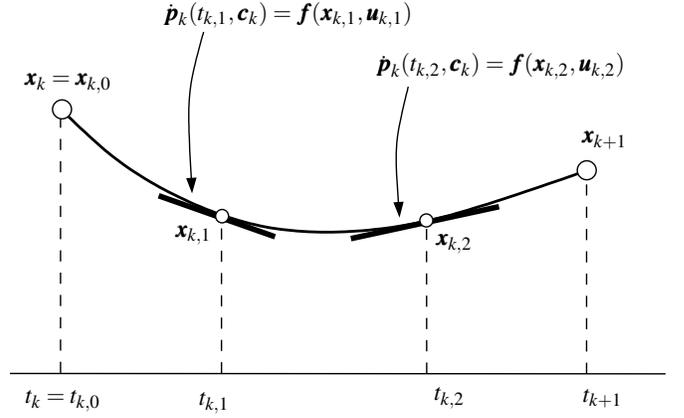}		
		{
			\psfrag{tk0}[l]{\small $t_{k}=t_{k,0}$}
			\psfrag{tk1}[l]{\small $t_{k,1}$}
			\psfrag{tk2}[l]{\small $t_{k,2}$}
			\psfrag{tkp}[l]{\small $t_{k+1}$}
			\psfrag{xk0}[l]{\small $\vr{x}_{k}=\vr{x}_{k,0}$}
			\psfrag{xk1}[l]{\small $\vr{x}_{k,1}$}
			\psfrag{xk2}[l]{\small $\vr{x}_{k,2}$}
			\psfrag{xend}[l]{\small $ \vr{x}_{k+1}$}%
		  \psfrag{xdot1}[l]{\small $\vr{\dot{p}}_k(t_{k,1},\vr{c}_k)=\vr{f}(\vr{x}_{k,1},\vr{u}_{k,1})$}
	    \psfrag{xdot2}[l]{\small $\vr{\dot{p}}_k(t_{k,2},\vr{c}_k)=\vr{f}(\vr{x}_{k,2},\vr{u}_{k,2})$}
		}
		\caption{In the interval $[t_k,t_{k+1}]$, the state trajectory $\vr{x}(t)$ is approximated by a polynomial of degree $d$ that interpolates $d+1$ states $\vr{x}_{k,0}, \ldots, \vr{x}_{k,d}$, where $\vr{x}_{k,0} = \vr{x}_k$ (the figure depicts the case $d=2$). The derivative of this polynomial must match the system dynamics at the collocation points $t_{k,1}, \ldots, t_{k,d}$. \label{fig:collocation}}
	\end{center}
\end{figure}

Transcription via Gauss-Legendre collocation works as follows. The form of
$\vr{x}(t)$ in the interval $\left[t_k, t_{k+1}\right]$ is not known a priori, but we assume it
to be approximated by a polynomial of degree $d$ taking the value $\vr{x}_k$ for
$t=t_k$. This polynomial is defined as the one that
interpolates $d+1$ states
\begin{equation} \nonumber
\vr{x}_{k,0}, \ldots, \vr{x}_{k,d},
\end{equation}
corresponding to $d+1$ time instants 
\begin{equation} \nonumber
t_{k,0}, \ldots, t_{k,d}
\end{equation}
in the interval $\left[t_k,t_{k+1}\right]$ (Fig.~\ref{fig:collocation}), where $t_{k,0}=t_k$, and
\begin{equation}
	t_{k,0} < t_{k,1} < \dots < t_{k,d}.
\end{equation}
Using Lagrange's interpolation formula~\cite{Berrut_SIAM2004}, we thus can write this polynomial as
\begin{equation} \label{eq:polynomial}
\vr{p}_k(t,\vr{c}_k) = \vr{x}_{k,0} \cdot \ell_{0}(t-t_k) + \dots + \vr{x}_{k,d} \cdot \ell_{d}(t-t_k),
\end{equation}
where 
\begin{equation}
	\vr{c}_k = (\vr{x}_{k,0},\ldots,\vr{x}_{k,d})
\end{equation}
is the vector of polynomial coefficients, 
and 
\begin{equation}
	\ell_{0}(t - t_k), \ldots, \ell_{d}(t-t_k)
\end{equation}
is the basis of Lagrange polynomials of degree $d$ defined for $t \in [t_k,t_{k+1}]$.
Recall that these polynomials only depend on $t_{k,0}, \ldots, t_{k,d}$, and
that
\begin{equation}
	\label{eq:lagrange_property}
	\ell_{j}(t_{k,i}-t_k) =
	\begin{cases}
		1, & \text{if}\ j=i \\
		0, & \text{otherwise}
	\end{cases}
\end{equation}
so $\vr{p}_k(t_{k,i},\vr{c}_k)=\vr{x}_{k,i}$ \cite{Berrut_SIAM2004}. Since we
wish $\vr{p}_k(t_k,\vr{c}_k) = \vr{x}_k$, it must be $\vr{x}_{k,0} = \vr{x}_k$
in Eq.~\eqref{eq:polynomial}, and we shall assume so hereafter. The remaining
coefficients $\vr{x}_{k,1},\ldots,\vr{x}_{k,d}$ are unknown, but they can be
determined by forcing $\vr{p}_k(t,\vr{c}_k)$ to satisfy
$\vr{\dot{x}}(t)=\vr{f}(\vr{x}(t),\vr{u}(t))$ for all $t = t_{k,1},\dots,t_{k,d}$. This
yields the~$d$ collocation constraints
\begin{equation}
	\label{eq:collocation}
	\vr{\dot{p}}_k(t_{k,i},\vr{c}_k) = 
	\vr{f}(\vr{x}_{k,i},\vr{u}_{k,i}), \quad 
	\quad i = 1,\dots, d,
\end{equation}
where $\vr{u}_{k,i} = \vr{u}(t_{k,i})$. The values $t_{k,1},\dots,t_{k,d}$ are
called the collocation points, and correspond to the roots of the Legendre
polynomial of degree $d$ \cite{Kelly_SIAM2017}. The left-hand side of Eq.
\eqref{eq:collocation}, in turn, is easy to formulate, as
$\vr{\dot{p}}_k(t_{k,i},\vr{c}_k)$ can be written as
\begin{equation}
	\vr{\dot{p}}_k(t_{k,i},\vr{c}_k)=\mt{J}(t_{k,0},\dots,t_{k,d}) \cdot \vr{c}_k,
\end{equation}
where $\mt{J}$ is a constant differentiation matrix that solely depends
on $t_{k,0},\dots,t_{k,d}$ \cite{Berrut_SIAM2004,Kelly_SIAM2017}. The
state $\vr{x}_{k+1}$ is finally given by
\begin{equation} 
	\label{eq:xkpCol}
	\vr{x}_{k+1} = \vr{p}_k(t_{k+1},\vr{c}_k).
\end{equation}

In each interval $[t_k,t_{k+1}]$, therefore, Eq.~\eqref{eq:OCP_dynamics} is
transcribed into Eqs.~\eqref{eq:collocation} and \eqref{eq:xkpCol}, where
$\vr{p}_k(t,\vr{c}_k)$ is defined by Eq.~\eqref{eq:polynomial}. Such a
transcription is very precise, as the local integration error of the Gauss-Legendre
 scheme is $O(\Delta t^{2d+1})$~\cite{Hairer_SPRINGER2006}. In the end,
once the transcribed problem is solved, the coefficients~$\vr{c}_k$ will be
known, and the trajectory $\vr{x}(t)$ will be the spline defined by
$\vr{p}_0(t,\vr{c}_0), \ldots, \vr{p}_{N-1}(t,\vr{c}_{N-1})$.

\section{Previous transcription methods} 
\label{sec:conventional_schemes}

We next review the three principal methods for transcribing Problem
\eqref{eq:OCP} given so far in the literature, for the case of constrained
mechanical systems (Sections \ref{subsec:basic_collocation} to
\ref{subsec:PKT}).

\subsection{Basic collocation} \label{subsec:basic_collocation}
A direct way of transcribing Problem \eqref{eq:OCP} consists in applying
the methods in Section \ref{sec:background} to each one of its equations. This
results in the optimization problem
\begin{mini!}[2]
	{\vr{w}} 						%
	{C(\vr{w})\label{eq:NLPcost}}	%
	{\label{eq:NLP}}				%
	{}								%
	\addConstraint
	{\vr{h}(\vr{x}_k,\vr{u}_k)}{\ge \vr{0},\quad\label{eq:NLPpath}}{\!k\!=\!0,\makebox[1.05em][c]{.\hfil.\hfil.},N,}
	\addConstraint
	{\vr{\dot{p}}_k(t_{k,i},\vr{c}_k)=}{\vr{f}({\vr{x}}_{k,i},\vr{u}_{k,i}),\quad\label{eq:NLPcol}}{\!k\!=\!0,\makebox[1.05em][c]{.\hfil.\hfil.},N\!-\!1,}
	\addConstraint
	{}{\nonumber}{\!i\!=\!1,\makebox[1.05em][c]{.\hfil.\hfil.},d,}
	\addConstraint
	{\vr{p}_k (t_{k+1},\vr{c}_k)=}{{\vr{x}}_{k+1},\quad\label{eq:NLPcont}}{\!k\!=\!0,\makebox[1.05em][c]{.\hfil.\hfil.},N\!-\!1,} 
	\addConstraint
	{\vr{b}(\vr{x}_0,\vr{x}_N)}{=\vr{0},\label{eq:NLPboundary}}{}
\end{mini!}
where $C(\vr{w})$ is the discrete version of the integral cost in
Eq.~\eqref{eq:OCP_cost}, and $\vr{w}$ encompasses all the decision variables
intervening: the actions $\vr{u}_0, \dots,\vr{u}_N$, the states
$\vr{x}_0,\dots,\vr{x}_{N}$, and the collocation states $\vr{x}_{k,i}$ for
$k=0,\dots,N-1$, $i=1,\dots,d$. This transcription will be called the basic
method hereafter.

The basic method is simple, but it presents an important limitation. As shown in
Fig.~\ref{fig:dae_methods}(a), the knot states $\vr{x}_0,\dots,\vr{x}_N$ tend to
drift off from $\X$ due to the discretization errors inherent to
Eqs.~\eqref{eq:NLPcol} and~\eqref{eq:NLPcont}, which leads to unrealistic
trajectories that can be difficult to control. Unfortunately, this problem
cannot be solved by just adding
\begin{equation}
	\vr{F}(\vr{x}_k) = \vr{0}, \quad k = 0, \ldots, N,
	\label{eq:extra_path}
\end{equation}
to the transcribed formulation, as Eqs.~\eqref{eq:NLPcol} and \eqref{eq:NLPcont}
already determine $\vr{x}_1,\dots,\vr{x}_N$ when $\vr{x}_0$ is known and a
control sequence is fixed. Therefore, the addition of \eqref{eq:extra_path}
would introduce redundant constraints, thus violating the linear independence
constraint qualification (LICQ) required by the Karush-Kuhn-Tucker conditions of
optimality~\cite{nocedal2006numerical}. Despite the problem of drift, however,
the simplicity of the basic method makes it a good tool to provide initial
guesses for more elaborate methods.

\subsection{Collocation with Baumgarte stabilization}
\label{subsec:baumgarte_stab}
An alternative to the basic method is to resort to Baumgarte stabilization. This
technique consists in altering the system dynamics by adding artificial forces
to damp the trajectory drift from $\X$. To achieve so, Eq.~\eqref{eq:dotdotphi}
is modified by adding stabilizing terms for the residuals of Eqs. \eqref{eq:phi}
and \eqref{eq:Bqdot}. The modified equation takes the form
\begin{equation}
	\label{eq:baumgarte}
	\vr{B}(\vr{q})\; \ddq +
	\alpha \; \mt{B}(\vr{q})\dq
	+
	\beta 
	\begin{bmatrix}
		\vr{\Phi}(\vr{q})\\
		\vr{0}_{n_r}
	\end{bmatrix}
	- \vr{\xi}(\vr{q},\dq)
	= \vr{0},
\end{equation}
where $\vr{0}_{n_r}$ is a column vector of $n_r$ zeros, and $\alpha$ and $\beta$
are constant paramaters that have to be tuned to ensure a stable trajectory near $\X$.
Following steps analogous to those in
Eqs.~\eqref{eq:dyn3}-\eqref{eq:explicitdynamics} we then obtain a stabilized
dynamics equation
\begin{equation}
	\label{eq:explicitdynamicsBaum}
	\vr{\dot{x}}=\vr{f}_\textrm{stab}(\vr{x},\vr{u}).
\end{equation}
Using Baumgarte stabilization, thus, the transcription in Problem \eqref{eq:NLP} is improved by replacing Eq.~\eqref{eq:NLPcol} by
\begin{equation}
	\label{eq:collocationBaumgarte}
	\vr{\dot{p}}_k(t_{k,i},\vr{c}_k) 
	= 
	\vr{f}_\textrm{stab}(\vr{x}_{k,i},\vr{u}_{k,i}) 
\end{equation}
for $k = 0, \ldots, N-1$ and $i=1,\ldots,d$.

The simplicity of the technique has made it a common approach to mitigate the
problem of drift, but the method presents three main drawbacks. First, the
tuning of $\alpha$ and $\beta$ is not trivial, as the dynamics from the
stabilizing terms should be faster than the drift dynamics produced by the
transcribed equations, otherwise the trajectory might depart from $\X$ in an
unstable manner. Despite the importance of this point, only heuristic rules
have been given to choose $\alpha$ and $\beta$ so far~\cite{Blajer_CMAME2011,rabier2002theoretical}.
Second, Eq.~\eqref{eq:explicitdynamicsBaum} has a more complex structure than
the one in Eq. \eqref{eq:explicitdynamics}, which complicates the computation of
gradients and Hessians needed by the optimizer. Third, and most notably, the
approach alters the dynamics of the system artificially, as the use of
Eq.~\eqref{eq:collocationBaumgarte}, instead of Eq.~\eqref{eq:NLPcol}, implies
that the actual dynamics will not be fulfilled at the collocation points. This
point is illustrated in Section \ref{sec:examples} with the help of examples.

\subsection{The PKT method}
\label{subsec:PKT}

In \cite{Posa_ICRA2016}, Posa et al. modify the classical Hermite-Simpson
collocation method \cite{hargraves1987direct} to deal with constrained systems.
In their case, the system can only have holonomic constraints (so $n_v = n_p$
and $d_\X = n_x - 2n_p$) and the trajectories are approximated by
polynomials of third degree. We next review this approach, which we refer to as the
PKT method.

As in \cite{hargraves1987direct}, the PKT method approximates the trajectory for
$t \in [t_k,t_{k+1}]$ by a cubic polynomial $\vr{p}_k(t,\vr{c}_k)$. This
polynomial interpolates $\vr{x}_k$ and $\vr{x}_{k+1}$ while satisfying the slopes
$\vr{\dot{x}}_k$ and $\vr{\dot{x}}_{k+1}$ determined by the
collocation constraints
\begin{subequations} 
	\label{eq:colHS}
	\begin{align}
		& \vr{\dot{x}}_k = \vr{f}(\vr{x}_{k},\vr{u}_{k}),
		\label{eq:colHS1} \\	
		& \vr{\dot{x}_{k+1}} = \vr{f}(\vr{x}_{k+1},\vr{u}_{k+1}).
		\label{eq:colHS2}
	\end{align}
\end{subequations}

In terms of $\vr{c}_k = (\vr{x}_{k},\vr{x}_{k+1})$, the polynomial $\vr{p}_k(t,\vr{c}_k)$ can be written as follows
\begin{align}
	\begin{split} \label{eq:HSinterpol}
		\vr{p}_k(t, & \vr{c}_k) = \vr{x}_k + \vr{f}_k (t-t_k) - \\
		& - \tfrac{(t-t_k)^2}{\Delta t^2}
		(3\vr{x}_k-3\vr{x}_{k+1}  + 
		2\Delta t \; \vr{f}_k + 
		\Delta t \; \vr{f}_{k+1}) + \\  
		& + \tfrac{(t-t_k)^3}{\Delta t^3}
		(2\vr{x}_k - 2\vr{x}_{k+1}  + 
		\Delta t \; \vr{f}_k + 
		\Delta t \; \vr{f}_{k+1}),
	\end{split}
\end{align}
where $\vr{f}_{k}=\vr{f}(\vr{x}_{k},\vr{u}_{k})$. To determine the value
of $\vr{x}_{k+1}$ in $\vr{p}_k(t, \vr{c}_k)$, a third collocation constraint is imposed at the midpoint $t_{k,c} = \tfrac{1}{2}(t_k + t_{k+1})$, which can be expressed
as
\begin{equation}
	\vr{\dot{p}}_k(t_{k,c},\vr{c}_k) = \vr{f}(\vr{x}_{k,c},\vr{u}_{k,c}),
	\label{eq:HSsingle} 
\end{equation}
where $\vr{x}_{k,c} = \vr{p}(t_{k,c},\vr{c}_k)$ and $\vr{u}_{k,c} =
\tfrac{1}{2}(\vr{u}_k+\vr{u}_{k+1})$. Since Eqs.~\eqref{eq:colHS} have already
been already been taken into account in Eq.~\eqref{eq:HSinterpol},
Eq.~\eqref{eq:HSsingle} is the sole constraint needed to transcribe the dynamics
in the usual Hermite-Simpson method \cite{hargraves1987direct}.

Posa et al. note in \cite{Posa_ICRA2016} that, while Eq.~\eqref{eq:HSsingle}
poses no challenge in unconstrained systems, it is problematic when holonomic
constraints are present. In the former case the $n_x$ components of
$\vr{x}_{k+1}$ are independent, so the $n_x$ conditions in~\eqref{eq:HSsingle}
properly determine $\vr{x}_{k+1}$. In the latter case, however, one must impose
\begin{equation}
	\label{eq:PKTxonX}
	\vr{F}(\vr{x}_k)=\vr{0}, \hspace{8mm} k=0,\ldots,N,
\end{equation}
to ensure all knot states lie on $\X$, so $\vr{x}_{k+1}$ will only have $d_\X =
n_x - 2 n_p$ freedoms and Eq.~\eqref{eq:HSsingle} will overconstrain its value. To
circumvent this problem, the PKT  method relaxes Eq.~\eqref{eq:HSsingle} to
include $2\,n_p$ new variables for each time interval, $\vr{\gamma}_{k,c}
\in \R{n_p}$ and $\vr{\lambda}_{k,c} \in \R{n_p}$, which provide the
required freedoms to eliminate the overconstraint. Specifically, 
\cite{Posa_ICRA2016} reformulates \eqref{eq:HSsingle} as
\begin{subequations} \label{eq:HS_PKT}
	\begin{align}
		\vr{\dot{q}}_k(t_{k,c},\vr{c}_k) &= \vr{v}_k(t_{k,c},\vr{c}_k) + \mt{B}(\vr{q}_{k,c})\tr \cdot \vr{\gamma}_{k,c}, 
		\label{eq:HS_PKT1} \\
		\vr{\dot{v}}_k(t_{k,c},\vr{c}_k) &= \vr{g}(\vr{x}_{k,c},\vr{u}_{k,c},\vr{\lambda}_{k,c}),
		\label{eq:HS_PKT2}
	\end{align}
\end{subequations}
where $\vr{q}_k(t,\vr{c}_k)$ and $\vr{v}_k(t,\vr{c}_k)$ refer to the configuration and velocity components of $\vr{p}_k(t,\vr{c}_k)$, and
\begin{equation}
	\label{eq:acceleration_function}
	\vr{g}(\vr{x},\vr{u},\vr{\lambda}) = 
	\mt{M}(\vr{q})^{-1} \cdot \left[\vr{\tau}(\vr{u},\vr{q},\vr{\dot{q}}) - \mt{B}(\vr{q})\tr \vr{\lambda}\right]
\end{equation}
is the acceleration function determined by Eq.~\eqref{eq:dynamics}. Under the
PKT method, therefore, the transcription in \eqref{eq:NLP} is improved
by adding the constraints in \eqref{eq:PKTxonX} and replacing \eqref{eq:NLPcol}
by \eqref{eq:HS_PKT} for $k=0,\ldots,N-1$.

The strong point of the PKT method is that it keeps the knot states $\vr{x}_0,
\ldots, \vr{x}_N$ on $\X$, while ensuring, through \eqref{eq:colHS}, that such
states satisfy the actual dynamics of the system. This is a clear improvement
over the Baumgarte method, which only keeps $\vr{x}_0, \ldots, \vr{x}_N$ near
$\X$ approximately, and violates the actual dynamics at the collocation points.
The main weakness of the PKT method, however, is that it modifies the system
dynamics at the midpoint of each interval. In particular,
Eq.~\eqref{eq:HS_PKT2}, relaxes the second order dynamics constraint, as it only
accounts for Lagrange's equation in \eqref{eq:dynamics} but not for the
acceleration constraint in \eqref{eq:dotdotphi}. This implies that, at each
midpoint $t_{k,c}$, the robot is treated as an unconstrained system subject to
the forces defined by the $\vr{\lambda}_{k,c}$ multipliers (see
\cite{Featherstone_book1987} which interprets these multipliers as external
forces applied at the breakpoints of kinematic loops). Thus, while the
collocation constraints at $t_0,\ldots,t_{N}$ are exactly enforced, those at the
midpoints $t_{0,c}, \ldots, t_{N-1,c}$ are not. Additional limitations of the
PKT method are that, as defined in \cite{Posa_ICRA2016}, it can only handle
holonomic systems, and it uses approximation polynomials of a fixed degree. The
latter, in particular, impedes its application in hp-adaptive meshing schemes.
  
\section{New transcription methods} \label{sec:new_schemes}

We next propose two new methods to solve the problem of drift. The two methods,
which are referred to as the projection and local coordinates methods, satisfy
the actual system dynamics at the collocation points, and they can use
approximation polynomials of arbitrary degree. While the projection method
guarantees that the knot states $\vr{x}_0, \ldots, \vr{x}_N$ lie on $\X$ exactly
(Section \ref{subsec:projection}), the local coordinates method  achieves full
drift elimination over the entire trajectory $\vr{x}(t)$ (Section
\ref{subsec:local}).

\subsection{The projection method} \label{subsec:projection}

\begin{figure}[t]
	\begin{center}
		\pstool[width=0.9\linewidth]{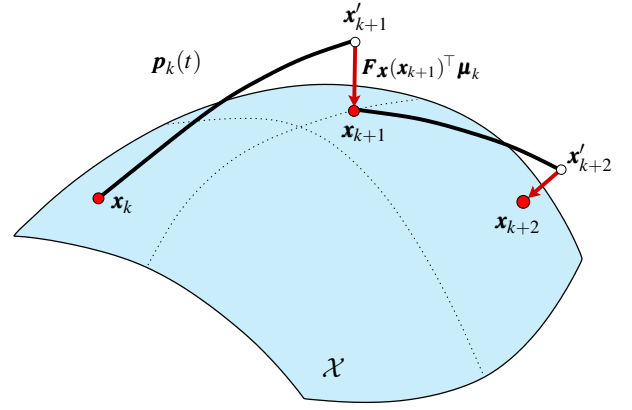}
		{
			\psfrag{X}[l]{$\X$}
			\psfrag{xk}[l]{\small $\vr{x}_k$}
			\psfrag{xkp}[l]{\small $\vr{x}_{k+1}$}
			\psfrag{xkp'}[l]{\small $\vr{x}_{k+1}'$}
			\psfrag{xkpp'}[l]{\small $\vr{x}_{k+2}'$}
			\psfrag{xkpp}[l]{\small $\vr{x}_{k+2}$}
			\psfrag{xt}[l]{\small $\vr{p}_k(t)$}
			\psfrag{d}[l]{\footnotesize $\mt{F}_{\vr{x}}(\vr{x}_{k+1})\tr\vr{\mu}_k$}
		}
	\end{center}
	\caption{\label{fig:projection} The projection method. The 
		end state of each polynomial is projected orthogonally to $\X$
		to eliminate the accumulated drift.}
\end{figure}

The projection method cancels the drift at each knot point by projecting the end
state of each Lagrange interpolation orthogonally to $\X$. The method only
requires modifying Problem~\eqref{eq:NLP} as follows. First, we add the constraints
\begin{equation}
	\label{eq:PROJxonX}
	\vr{F}(\vr{x}_k)=\vr{0}, \hspace{8mm} k=0,\ldots,N,
\end{equation}
to ensure all knot states $\vr{x}_0, \ldots, \vr{x}_N$ lie on $\X$. Second, we add~$N$
auxiliary states
\begin{equation} 
	\nonumber
	\vr{x}_{1}', ..., \vr{x}_{N}',
\end{equation}
and $n_e$-dimensional vectors
\begin{equation} 
	\nonumber
	\vr{\mu}_{1}, ..., \vr{\mu}_{N},
\end{equation}
to the decision variables $\vr{w}$. Finally, we replace Eq.~\eqref{eq:NLPcont} by
\begin{equation}
	\vr{x}_{k+1}'=\vr{p}_k(t_{k+1},\vr{c}_k) \quad \quad k=0,\dots,N-1,
\end{equation}
so $\vr{x}_{k+1}'$ is the end state of the Lagrange interpolation, and
add the constraint
\begin{align}
\label{eq:muk}
		\vr{x}_{k+1}=\vr{x}_{k+1}' + 
		\mt{F}_\vr{x}(\vr{x}_{k+1})\tr \vr{\mu}_k \quad \quad &k=0,\dots,N-1
\end{align}
to ensure that $\vr{x}_{k+1}$ is the orthogonal projection of $\vr{x}_{k+1}'$ on
$\X$ (Fig. \ref{fig:projection}). 

Note that the rows of $\mt{F}_\vr{x}$ provide a basis of the normal space of
$\X$ at $\vr{x}_{k+1}$, so $\mt{F}_\vr{x}(\vr{x}_{k+1})\tr \vr{\mu}_k$ will be a
normal vector of $\X$ at $\vr{x}_{k+1}$. Thus, although $\vr{x}_{1}', ...,
\vr{x}_{N}'$ may deviate from $\X$, the drift will not accumulate because the
joint effect of Eqs.~\eqref{eq:PROJxonX} and~\eqref{eq:muk} will fully remove it
after every time step. Actually, the local integration error before the
projection is $O(\Delta t^{2d+1})$, and Hairer shows that the projection step
does not affect this bound \cite{Hairer_SPRINGER2006,hairer1996solving}.

Regarding the problem of drift, both the projection and the PKT methods achive a
similar result, since both ensure $\vr{x}_{1}, ..., \vr{x}_{N}$ lie on $\X$
exactly. However, the projection method is advantageous in that it fulfills the
actual dynamics at all collocation points, and its approximation polynomial,
$\vr{p}_k(t,\vr{c}_k)$, can be of arbitrary degree. The weak point of the
projection method is that, as in the PKT method, some drift from~$\X$ will
persist despite the projections. If a driftless continuous trajectory is
required, however, we can always resort to the following method.

\subsection{The local coordinates method} \label{subsec:local}

The idea of this method is to transcribe Eq. \eqref{eq:OCP_dynamics} by using
integration in local coordinates
\cite{Potra_JSM1991,Hairer_SPRINGER2006,potra1991numerical}. In trajectory
optimization, this concept was applied to $SO(3)$ and multiple shooting in
\cite{Manara_ND2017}, and we here extend it to general manifolds using
collocation methods. We first explain the idea using generic maps
(Section~\ref{subsec:local}.1) and then give the collocation
equations for particular choices of such maps (Section~\ref{subsec:local}.2).\\[-0.1cm]

\noindent{\em B.1. Collocation in local coordinates}\\[-0.25cm]

\begin{figure}[t!]
	\begin{center}
		\pstool[width=0.9\linewidth]{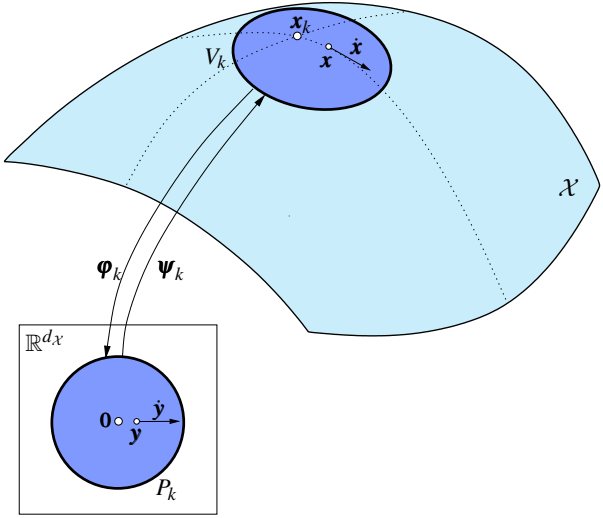}
		{
			\psfrag{X}[l]{\small $\X$}
			\psfrag{(a)}[l]{\small (a)}
			\psfrag{(b)}[l]{\small (b)}
			\psfrag{x}[l]{\small $\vr{x}$}
			\psfrag{y}[l]{\small $\vr{y}$}
			\psfrag{0}[l]{\footnotesize $\vr{0}$}
			\psfrag{xdot}[l]{\small $\vr{\dot{x}}$}
			\psfrag{ydot}[l]{\small $\dot{\vr{y}}$}
			\psfrag{R}[l]{\small $\R{d_{\X}}$}
			\psfrag{Ui}[l]{\small $P_k$}
			\psfrag{Vi}[l]{\small $V_k$}
			\psfrag{psii}[l]{\small $\vr{\psi}_k$}
			\psfrag{phii}[l]{\small $\vr{\varphi}_k$}
			\psfrag{xc}[l]{\small $\vr{x}_k$}
		}
	\end{center}
	\caption{\label{fig:charts} A chart of $\X$ is a map $\vr{\varphi}_k$
		from an open neighbourhood $V_k$ of $\X$ to an open neighbourhood $P_k$ of $\R{d_\X}$. 
		The inverse map $\vr{\psi}_k$ gives a local parametrization of $V_k \subset \X$.}
\end{figure}

Let $\vr{y} = \vr{\varphi}_k(\vr{x})$ be a local chart
of $\X$ defined at
some point $\vr{x}_k \in \X$. This implies that $\vr{\varphi}_k$ is a local diffeomorphism 
\begin{equation} 
	\nonumber
	\vr{\varphi}_k : V_k \rightarrow P_k,
\end{equation}
where $V_k$ and $P_k$ are open neighbourhoods in $\X$ and $\R{d_\X}$,
respectively, including $\vr{x}_k$ and $\vr{\varphi}_k(\vr{x}_k)$. Without loss
of generality, we will assume that $\vr{\varphi}_k(\vr{x}_k)=\vr{0}$, so
$\vr{\varphi}_k$ maps $V_k$ to a neighborhood of the origin of
$\R{d_\X}$~(Fig.~\ref{fig:charts}). Since~$\vr{\varphi}_k$ is a diffeomorphism,
it has an inverse map
\begin{equation} 
	\nonumber
	\vr{\psi}_k : P_k \rightarrow V_k,
\end{equation}
which provides a local parametrization of $V_k \subset \X$. The fact that $\X$
is a smooth manifold guarantees that both~$\vr{\varphi}_k$ and~$\vr{\psi}_k$
exist for any point $\vr{x}_k\in\X$ \cite{Lee_2001}.

\begin{figure}[t!]
	\begin{center}
		\pstool[width=.9\linewidth]{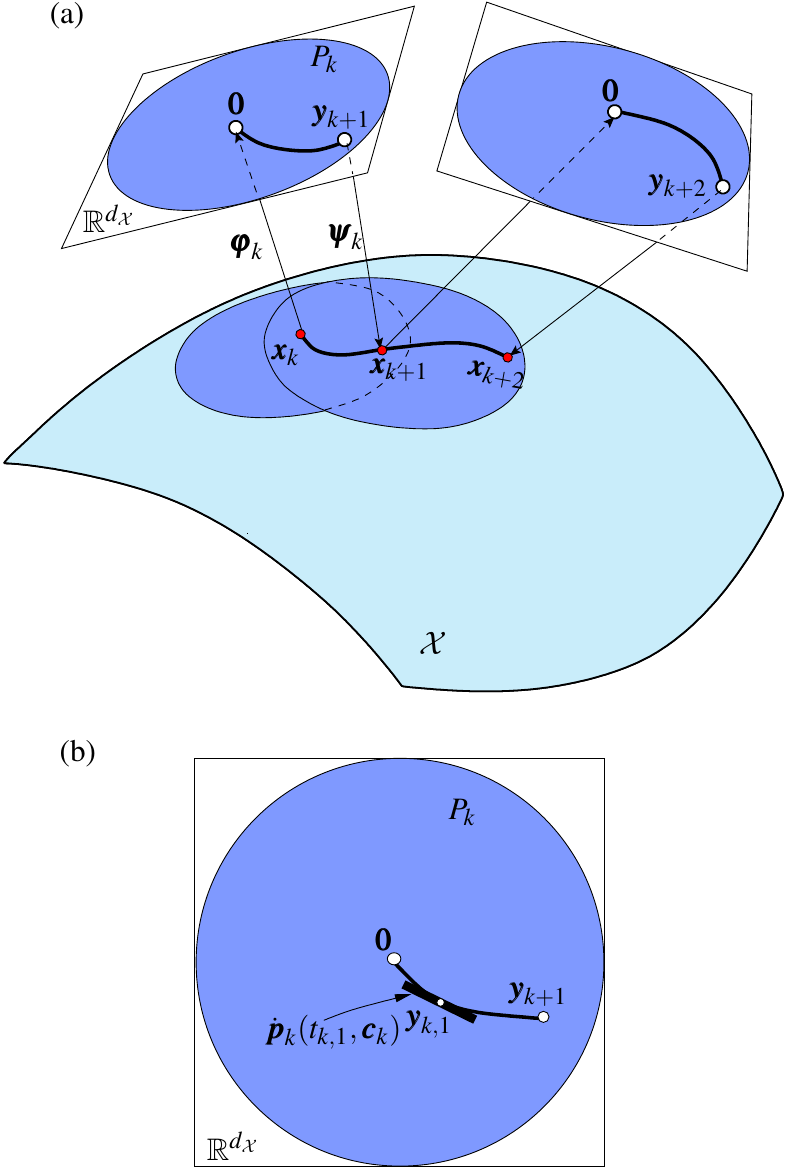}
		{
			\psfrag{a}[l]{\small (a)}
			\psfrag{b}[l]{\small (b)}
			\psfrag{Pk}[l]{\small $P_k$}	
			\psfrag{RdX}[l]{\small $\R{d_\X}$}
			\psfrag{X}[l]{\small $\X$}
			\psfrag{x1}[l]{\small $\vr{x}_{1}$}
			\psfrag{xk}[l]{\small $\vr{x}_{k}$}
			\psfrag{xkp}[l]{\small $\vr{x}_{k+1}$}
			\psfrag{xkpp}[l]{\small $\vr{x}_{k+2}$}
			\psfrag{xn}[l]{\small $\vr{x}_{N}$}
			\psfrag{yk}[l]{\small $\vr{0}$}		
			\psfrag{ykp}[l]{\small $\vr{0}$}		
			\psfrag{yk1}[l]{\small $\vr{y}_{k,1}$}
			\psfrag{yend}[l]{\small $\vr{y}_{\scriptsize k+1}$}
			\psfrag{yendp}[l]{\small $\vr{y}_{\scriptsize k+2}$}
			\psfrag{ydot1}[l]{\small $\vr{\dot{p}}_k(t_{k,1},\vr{c}_k)$}
			\psfrag{psik}[l]{\small $\vr{\psi}_{k}$}
			\psfrag{psikp}[l]{\small $\vr{\psi}_{k+1}$}
			\psfrag{phi}[l]{\small $\vr{\varphi}_{k}$}	
			\psfrag{phikp}[l]{\small $\vr{\varphi}_{k+1}$}
			\psfrag{T1}[l]{\small $P_k$}
		}
		\caption{Transcription using local coordinates.
			(a) To obtain $\vr{x}_{k+1}$, we map $\vr{x}_k$ 
			to $\vr{\varphi}_k(\vr{x}_k)=\vr{0} \in \R{d_\X}$, then integrate Eq.~\eqref{eq:localDyn4}
			to find $\vr{y}_{k+1}$, and finally project $\vr{y}_{k+1}$ to $\X$
			using $\vr{\psi}_k$.
			(b) The collocation constraint is now enforced in 
			$\R{d_\X}$. In the figure, we assume $d=1$, so
			$\vr{\dot{p}}_k(t_{k,1})$ must match $\vr{g}_k(\vr{y}_{k,1},\vr{u}(t_{k,1}))$.
			\label{fig:localCollocation} }
	\end{center}
\end{figure}

We next see that, using $\vr{\varphi}_k$ and~$\vr{\psi}_k$, we can compute the
state $\vr{x}_{k+1}$ that would be reached from $\vr{x}_k$ while ensuring that
$\vr{F}(\vr{x}_{k+1})=\vr{0}$ precisely. To this end, we first take the time
derivative of
\begin{equation}
	\vr{y} = \vr{\varphi}_k(\vr{x})
\end{equation}
and obtain 
\begin{equation} 
	\label{eq:localDyn1}
	\vr{\dot{y}} = \mt{S}(\vr{x}) \cdot \vr{\dot{x}},
\end{equation}
where $\mt{S}(\vr{x})=\partial\vr{\varphi}_k(\vr{x})/\partial \vr{x}$. If we
substitute $\vr{\dot{x}}=\vr{f}(\vr{x},\vr{u})$ in Eq.~\eqref{eq:localDyn1}, we
get
\begin{equation} 
	\label{eq:localDyn2} 
	\vr{\dot{y}} = \mt{S}(\vr{x}) \cdot \vr{f}(\vr{x},\vr{u}),
\end{equation}
and using $\vr{x} = \vr{\psi}_k(\vr{y})$ we arrive at
\begin{equation} 
	\label{eq:localDyn3}
	\vr{\dot{y}} = \mt{S}( \vr{\psi}_k(\vr{y})) \cdot \vr{f}(\vr{\psi}_k(\vr{y}),\vr{u}),
\end{equation}
which we compactly write as
\begin{equation} 
	\label{eq:localDyn4}
	\vr{\dot{y}} = \vr{g}_k(\vr{y},\vr{u}).
\end{equation}
Eq. \eqref{eq:localDyn4} provides the dynamic equation in $\vr{y}$ coordinates,
and we can use it as follows to compute
$\vr{x}_{k+1}$~[Fig.~\ref{fig:localCollocation}(a)]: using $\vr{\varphi}_k$, we
first map $\vr{x}_k$ to the origin of $\R{d_\X}$; we then integrate
Eq.~\eqref{eq:localDyn4} to find the state \mbox{$\vr{y}(t_{k+1}) =
	\vr{y}_{k+1}$} that would be reached from $\vr{y}(t_k)=\vr{0}$ under the actions
$\vr{u}(t)$; and, finally, we project $\vr{y}_{k+1}$ back to $\X$ using
\begin{equation}
	\label{eq:yCont}
	\vr{x}_{k+1} = \vr{\psi}_k(\vr{y}_{k+1}).  
\end{equation}
Using this process, it is clear that $\vr{x}_{k+1}\in \X$ by construction.

To integrate Eq.~\eqref{eq:localDyn4} using collocation, we assume that the
trajectory $\vr{y}(t)$ is well approximated by a polynomial that starts at the
origin of $\R{d_\X}$ and interpolates $d$ unknown states
\begin{equation*}
	\vr{y}_{k,1},\dots,\vr{y}_{k,d}
\end{equation*}
corresponding to the Gauss-Legendre collocation times
\begin{equation}
	t_{k,1}<\dots<t_{k,d}
\end{equation}
from $\left[t_k,t_{k+1}\right]$ [Fig. \ref{fig:localCollocation}(b)]. This polynomial can be written as
\begin{equation} \label{eq:yPoly}
	\vr{p}_k(t,\vr{c}_k) = \vr{y}_{k,1} \cdot \ell_{1}(t-t_k) + \dots + \vr{y}_{k,d} \cdot \ell_{d}(t-t_k),
\end{equation}
where now
\begin{equation}
	\vr{c}_k = ( \vr{y}_{k,1}, \ldots,  \vr{y}_{k,d} ).
\end{equation}
Note that the polynomial in Eq. \eqref{eq:yPoly} satisfies
$\vr{p}_k(t_k,\vr{c}_k)=\vr{0}$ as required since, due to Eq.
\eqref{eq:lagrange_property}, $\ell_{1}(t-t_k),\dots,\ell_{d}(t-t_k)$ are all
zero for $t=t_k$. To determine $\vr{y}_{k,1},\dots,\vr{y}_{k,d}$, we only have
to impose the $d$ collocation constraints
\begin{equation}
	\vr{\dot{p}}_k(t_{k,i},\vr{c}_k) =\vr{g}_k(\vr{y}_{k,i}, \vr{u}_{k,i}), \quad \textrm{for} \quad  i = 1, \dots, d.
	\label{eq:yCol}
\end{equation}
The value of $\vr{y}_{k+1}$ will then be given by
\begin{equation}
	\vr{y}_{k+1} = \vr{p}_k(t_{k+1},\vr{c}_k),
\end{equation}
so Eq. \eqref{eq:yCont} can be written as
\begin{equation}
	\vr{x}_{k+1} = 
	\vr{\psi}_k(\vr{p}_k(t_{k+1},\vr{c}_k)).
\end{equation}

With the earlier procedure, thus, we can achieve a drift-free transcription of
Problem \eqref{eq:OCP} by replacing Eqs.~\eqref{eq:NLPcol}
and~\eqref{eq:NLPcont} by
\begin{subequations}
	\begin{align}
		\vr{\dot{p}}_k(t_{k,i},\vr{c}_k) = \vr{g}_k(\vr{y}_{k,i},\vr{u}_{k,i}), \quad &k=0,\dots,N-1, \label{eq:NLP_local1} \\	
		&i= 1, \dots, d, \nonumber \\
		\vr{x}_{k+1} = \vr{\psi}_k(\vr{p}_k(t_{k+1},\vr{c}_k)), \quad &k=0,\dots,N-1.  \label{eq:NLP_local2}
	\end{align}
\end{subequations}

Once the transcribed problem is solved we can use $\vr{p}_0(t,\vr{c}_0),\dots,
\vr{p}_{N-1}(t,\vr{c}_{N-1})$ to construct a continuous spline for $\vr{x}(t)$
lying in $\X$ for all $t$, something that the earlier methods were unable to
obtain. This spline will be given by
\begin{equation}
	\vr{\psi}_0(\vr{p}_0(t,\vr{c}_0)),\dots,\vr{\psi}_{N-1}
	(\vr{p}_{N-1}(t,\vr{c}_{N-1})).
\end{equation}

\vspace{2mm}
\noindent{\em B.2. Collocation in tangent space coordinates}\\[-0.25cm]

\begin{figure}[t]
	\begin{center}
		\pstool[width=0.87\linewidth]{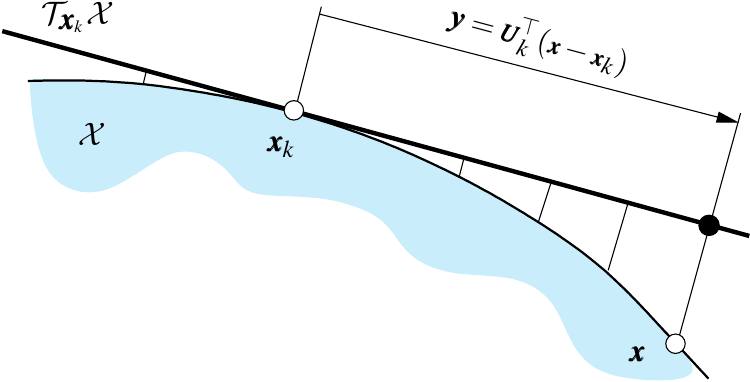}
		{
			\psfrag{X}[l]{\small $\X$}
			\psfrag{x}[l]{\small $\vr{x}$}
			\psfrag{xz}[l]{\small $\vr{x}$}
			\psfrag{T}[l]{\small $\TX{\vr{x}_k}$} 
			\psfrag{yvec}[l]{\small $\vr{y} = {\scriptsize \mt{U}_k\tr(\vr{x}-\vr{x}_k)}$}
			\psfrag{x0}[l]{\small $\vr{x}_k$}
		}
	\end{center}
	\caption{\label{fig:chartsTangent}
		The tangent space parametrization. The map $\vr{\varphi}_k$ is defined by 
		the projection of $\vr{x}$ onto the tangent space $\TX{\vr{x}_k}$.}
\end{figure}

For some manifolds, the maps $\vr{\psi}_k$ and $\vr{\varphi}_k$ can be defined
in closed form (for example, by expressing some variables of Eq.~\eqref{eq:F} as
a function of the others). For the sake of generality, however, we here define
them using tangent space coordinates, which work for any manifold
\cite{Potra_JSM1991,Hairer_SPRINGER2006}. Using these coordinates, the map
$\vr{y} = \vr{\varphi}_k(\vr{x})$ is obtained by projecting $\vr{x}$
orthogonally onto $\TX{\vr{x}_k}=\R{d_\X}$, as shown in Fig
\ref{fig:chartsTangent}, and takes the form
\begin{equation}\label{eq:phic}
\vr{y}=\mt{U}_k(\vr{x}_k)\tr \; (\vr{x}-\vr{x}_{k}),
\end{equation}
where $\mt{U}_k(\vr{x}_k)$ is an $n_x \times d_\X$ matrix whose columns provide
an orthonormal basis of $\TX{\vr{x}_k}$ (see Appendix~\ref{app:tangent_basis}).
The inverse map \mbox{$\vr{x}=\vr{\psi}_k(\vr{y})$} is implicitly determined by
the system of nonlinear equations
\begin{subequations}
	\begin{empheq}[left=\empheqlbrace]{align}
		&\vr{y}-\mt{U}_k(\vr{x}_k)\tr(\vr{x}-\vr{x}_k) = \vr{0},\label{eq:TS_param1}\\
		&\vr{F}(\vr{x}) = \vr{0}\label{eq:TS_param2},
	\end{empheq}
\end{subequations}
which will be written as
\begin{equation}
	\label{eq:psiImpl}
	\vr{G}_k(\vr{x},\vr{y}) = \vr{0}
\end{equation}
for short. The time derivative of Eq. \eqref{eq:phic} provides the particular
form of Eq. \eqref{eq:localDyn1},
\begin{equation}
	\vr{\dot{y}} = \mt{U}_k(\vr{x}_k)\tr \; \vr{\dot{x}},
\end{equation}
where $\mt{U}_k(\vr{x}_k)\tr$ corresponds to $ \mt{S}(\vr{x})$. Eq.
\eqref{eq:localDyn2} is then given by
\begin{equation}
	\vr{\dot{y}} = \mt{U}_k(\vr{x}_k)\tr \; \vr{f}(\vr{x},\vr{u}).
\end{equation}

Since the $\vr{\psi}_k$  map is only defined implicitly by Eq.
\eqref{eq:psiImpl}, we cannot obtain Eq. \eqref{eq:localDyn3} explicitly using
these coordinates. However, using Eq. \eqref{eq:psiImpl}, we can still impose
Eq.~\eqref{eq:NLP_local1} via
\begin{subequations}
	\begin{align} 
	\label{eq:NLP_local11}
	\vr{G}_k(\vr{x}_{k,i},\vr{y}_{k,i}) = \vr{0}, \quad &k=0,\dots,N-1, \\
	&i=1,\dots,d, \nonumber \\
	\label{eq:NLP_local12}
	\vr{\dot{p}}_k(t_{k,i},\vr{c}_k) = \mt{U}_k(\vr{x}_k)\tr \, \vr{f}(\vr{x}_{k,i},\vr{u}_{k,i}), \quad
	&k=0,\dots,N-1, \\
	&i = 1, \dots, d, \nonumber
\end{align}
\end{subequations}
and Eq.~\eqref{eq:NLP_local2} via
\begin{equation}
		\label{eq:NLP_local22}
		\vr{G}_k(\vr{x}_{k+1},\vr{p}_k(t_{k+1},\vr{c}_k)) = \vr{0}, \quad k=0,\dots,N-1,
\end{equation}
where $\vr{y}_{k,1},\dots,\vr{y}_{k,d}$ must be added to the decision variables
$\vr{w}$ of the problem. 

In~\cite{Bordalba_TRO2020}, constraints are considered
that help to delimit the domains in which the local parameterizations are valid.
These constraints could be added to the transcription if necessary, but we omit
them to simplify the presentation, and because they are seldom useful when the
mesh of knot points is dense enough.

The local error of this method is the one of Gauss-Legendre collocation, i.e.
$O(\Delta t^{2d+1})$ (Section \ref{subsec:trans_diff_constraint}), but in
tangent space. For multi-step methods it has been proved that, under mild
conditions, the mapping back to the manifold does not increase the order of this
error~\cite{yen1993constrained}. This result probably also holds in general, and
in fact is taken for granted in \cite{rabier2002theoretical}, but providing a
formal proof of this point is left out of the scope of this paper.

\section{Implementation}
\label{sec:implementation_details}

To compare the techniques in Section \ref{sec:new_schemes} with those in Section
\ref{sec:conventional_schemes}, we have implemented them using MATLAB and the
symbolic toolbox for nonlinear optimization and algorithmic differentiation
CasADi~\cite{CasADi}. CasADi provides the necessary means to formulate the
problems and to compute the gradients and Hessians of the transcribed equations
using automatic differentiation. These derivatives are necessary to solve the
NLP problems that result, a task for which we rely on the
interior-point solver IPOPT \cite{IPOPT} in conjunction with the linear solver
MA-27~\cite{hsl2007}. Our implementation can be accessed 
through~\cite{bordalba2022git}. 

We next discuss important aspects that must be
considered when implementing and solving the transcribed problems, regardless of
the software platform employed.

\begin{table*}
	\caption{Summary of the variables and equations introduced by the five methods using the implicit formulation} \label{tab:counting}
	\begin{center}
		\begin{small}
			\setlength{\tabcolsep}{10pt}
			\begin{tabular}{lllr}
				\toprule
				{\bf Method} &
				{\bf Variables / Equations} &
				\multicolumn{1}{l}{\bf Range} &
				\multicolumn{1}{r}{\bf Size}  \\
				\midrule
				\multirow{5}{*}{Basic and Baumgarte}&
				$\vr{u}_k$, $\vr{x}_k$   &
				$k=0,\ldots,N$ &
				$(n_u + 2 \: n_q)\:(N+1)$       \\ 	
				& 
				$\vr{x}_{k,i}$, $\vr{\ddot{q}}_{k,i}$, $\vr{\lambda}_{k,i}$     &
			  $k=0,\ldots,N-1;\; i=1,\ldots,d$ &
				$(3 \: n_q + \: n_v)\:N \: d$ \\ 
				& & & \\[-0.2cm]
				\cline{2-4}
				& & & \\[-0.2cm]
				& %
				$\vr{F}(\vr{x}_0)=\vr{0}$   &
				&  %
				$n_e$\\
				& %
				$\vr{p}_k (t_{k+1},\vr{c}_k)={\vr{x}}_{k+1}$  & 
				$k=0,\ldots,N-1$  & 
				$2 \: n_q\:N$ \\
				& %
				$\vr{\dot{p}}_k(t_{k,i},\vr{c}_k)=\vr{\dot x}_{k,i}$ &
				$k=0,\ldots,N-1; \; i=1,\ldots,d$ & 
				$2 \: n_q \: N\:d$   \\
				& %
				$\vr{D}(\vr{x}_{k,i},\vr{\dot{x}}_{k,i},\vr{u}_{k,i},\vr{\lambda}_{k,i}) = \vr{0}$ &
				$k=0,\ldots,N-1; \; i=1,\ldots,d$ &
				$(n_q + n_v)\:N\:d$    \\	%
				\midrule
				\multirow{5}{*}{PKT}   & 
				$\vr{u}_k$, $\vr{x}_k$, $\vr{\lambda}_{k}$ & 
				$k=0,\ldots,N$  & 
				$(n_u + 2 \: n_q+ n_p)\:(N+1)$       \\ 	
				&  %
				$\vr{\lambda}_{k,c}$, $\vr{\gamma}_{k,c}$  & 
				$k=0,\ldots,N-1$ & 
				$2\: n_p \: N$  \\ 
				& & & \\[-0.2cm] 
				\cline{2-4}
				& & & \\[-0.2cm]
				& %
				$\vr{F}(\vr{x}_k)=\vr{0}$ &
				$k=0,\ldots,N$ & 
				$2 \: n_p \: (N+1) $\\
				& %
				$\vr{B}(\vr{q}_k)\; \vr{g}(\vr{x}_k,\vr{u}_k,\vr{\lambda}_k) = \vr{\xi}(\vr{q}_k,\vr{\dot q}_k)$ &
				$k=0,\ldots,N$ &
				$n_p \: (N+1)$
				\\
				& %
				$\vr{\dot{q}}_k(t_{k,c},\vr{c}_k) = \vr{v}_k(t_{k,c},\vr{c}_k) + \mt{B}(\vr{q}_{k,c})\tr \: \vr{\gamma}_{k,c}$ &
				$k=0,\ldots,N-1$ & 
				$n_q \: N$  \\ 
				&  %
				$\vr{\dot{v}}_k(t_{k,c},\vr{c}_k) = \vr{g}(\vr{x}_{k,c},\vr{u}_{k,c},\vr{\lambda}_{k,c})$&
				$k=0,\ldots,N-1$  &
				$n_q \: N$ \\	
				\midrule
				\multirow{7}{*}{Projection} &
				$\vr{u}_k$, $\vr{x}_k$ &
				$k=0,\ldots,N$  & 
				$(n_u + 2\: n_q) \: (N+1)$ \\
				& %
				$\vr{x}_k'$, $\vr{\mu}_k$ &
				$k=1,\ldots,N$  & 
				$(2 \: n_q + n_e) \: N$ \\
				& %
				$\vr{x}_{k,i}$, $\vr{\ddot{q}}_{k,i},\vr{\lambda}_{k,i}$ &
				$k=0,\ldots,N-1; \; i=1,\ldots,d$ & 
				$(3\:n_q+n_v) \: N \: d$  \\
				& & & \\[-0.2cm] 
				\cline{2-4}
				& & & \\[-0.2cm]
				& %
				$\vr{F}(\vr{x}_{k}) = \vr{0}$  &
				$k=0,\ldots,N$ &
				$n_e \: (N+1)$\\
				& %
				$\vr{x}_{k+1}'=\vr{p}_k(t_{k+1},\vr{c}_k)$ &
				$k=0,\ldots,N-1$ & 
				$2 \: n_q \: N$  \\ 
				&  %
				$\vr{x}_{k+1}=\vr{x}_{k+1}'+\mt{F}_\vr{x}(\vr{x}_{k+1})\tr \vr{\mu}_k$   &
				$k=0,\ldots,N-1$ &  
				$2 \: n_q \: N$   \\	
				& %
				$\vr{\dot{p}}_k(t_{k,i},\vr{c}_k)=\vr{\dot x}_{k,i} $ &
				$k=0,\ldots,N-1; \; i=1,\ldots,d$ & 
				$2 \: n_q \: N \: d$  \\ 
				&  %
				$\vr{D}(\vr{x}_{k,i},\vr{\dot{x}}_{k,i},\vr{u}_{k,i},\vr{\lambda}_{k,i}) = \vr{0}$
				&  
				$k=0,\ldots,N-1; \; i=1,\ldots,d$
				&   
				$(n_q + n_v) \: N \: d $ \\	
				\midrule 
				\multirow{5}{*}{Local coordinates} &
				$\vr{u}_k$, $\vr{x}_k$ &
				$k=0,\ldots,N$  & 
				$(n_u + 2\: n_q) \: (N+1)$ \\
				& %
				$\vr{x}_{k,i}$, $\vr{y}_{k,i}$, $\vr{\ddot{q}}_{k,i}$, $\vr{\lambda}_{k,i}$  &
				$k=0,\ldots,N-1; \; i=1,\ldots,d$ &
			  $(3\:n_q+d_\X+n_v) \: N \: d$ \\
				& & & \\[-0.2cm] 
				\cline{2-4}
				& & & \\[-0.2cm]
				& %
				$\vr{F}(\vr{x}_0)=\vr{0}$ &
				&  %
				$n_e$\\
				& %
				$\vr{G}_k(\vr{x}_{k+1},\vr{p}_k(t_{k+1},\vr{c}_k)) = \vr{0}$ &
				$k=0,\ldots,N-1$ &
				$2 \: n_q \: N$  \\
				& %
				$\vr{G}_k(\vr{x}_{k,i},\vr{y}_{k,i}) = \vr{0}$ & 
				$k=0,\ldots,N-1; \;i=1,\ldots,d$  & 
				$2 \:n_q \: N \: d $ \\
				&  %
				$\vr{\dot{p}}_k(t_{k,i},\vr{c}_k) = \mt{U}_k(\vr{x}_k)\tr \, \vr{\dot x}_{k,i}$				
				&  
				$k=0,\ldots,N-1; \;i=1,\ldots,d$ &
				$d_\X \: N \: d $\\	
				& %
				$\vr{D}(\vr{x}_{k,i},\vr{\dot{x}}_{k,i},\vr{u}_{k,i},\vr{\lambda}_{k,i}) = \vr{0}$ & 
				$k=0,\ldots,N-1; \;i=1,\ldots,d$  &
				$(n_q + n_v) \: N \: d$
				\\	
				\bottomrule
			\end{tabular}		
		\end{small}	
	\end{center}
\end{table*}

\subsection{Explicit versus implicit dynamics}
\label{subsec:implicit_vs_explicit} 

In all transcriptions so far, the collocation constraints have been formulated
using the explicit form of the system dynamics in
Eq.~\eqref{eq:explicitdynamics}. The derivation of
Eq.~\eqref{eq:explicitdynamics}, however, requires finding the matrix inverse in Eq.~\eqref{eq:dyn4}, which often complicates the expressions of
the gradients and Hessians needed by the optimizer. Unless such an inverse is
simple enough, it may be preferable to write the collocation constraints using the
implicit form of the dynamics given by Eq.~\eqref{eq:dyn3}. This is easy to do, as 
it only requires the substitution of each term $\vr{f}(\vr{x}_{k,i},\vr{u}_{k,i})$ 
appearing in a collocation constraint by a new variable $\vr{\dot{x}}_{k,i}$ subject to
\begin{equation}
	\label{eq:implicit_dynamics}
	\vr{D}(\vr{x}_{k,i},\vr{\dot{x}}_{k,i},\vr{u}_{k,i},\vr{\lambda}_{k,i}) = \vr{0},
\end{equation}
where $\vr{D}(\vr{x},\vr{\dot{x}},\vr{u},\vr{\lambda}) = \vr{0}$ denotes
Eq.~\eqref{eq:dyn3}, or its stabilized version in the Baumgarte method, and
$\vr{\lambda}_{k,i}$ is an auxiliary multiplier. Clearly, this adds more
variables and equations to the transcribed problems, but the resulting system is
much sparser, which often improves the convergence of the optimizer. Moreover,
bounds on the constraint forces are easier to set in this form, as they directly
relate to the $\vr{\lambda}_{k,i}$ values. To set such bounds, however, the
$\vr{\ddot{q}}_{k,i}$ components of $\vr{\dot{x}}_{k,i}$ are not necessary, so
some approaches, and in particular the PKT method, use a semi-implicit approach
to eliminate them. This can be achieved by substituting each variable
$\vr{\ddot{q}}_{k,i}$ of the previous formulation by
$\vr{g}(\vr{x}_{k,i},\vr{u}_{k,i},\vr{\lambda}_{k,i})$, where
$\vr{g}(\vr{x},\vr{u},\vr{\lambda})$ is the acceleration function we obtained
in Eq.~\eqref{eq:acceleration_function}, and adding Eq.~\eqref{eq:dotdotphi} evaluated for
$t = t_{k,i}$, which yields
\begin{equation}
	\vr{B}(\vr{q}_{k,i})\; \vr{g}(\vr{x}_{k,i},\vr{u}_{k,i},\vr{\lambda}_{k,i}) = \vr{\xi}(\vr{q}_{k,i},\vr{\dot{q}}_{k,i}).
\end{equation}
This formulation reduces the number of added variables, but also the sparsity of
the system, so it may be less effective than the fully implicit formulation in our experience.

Table~\ref{tab:counting} summarizes the variables and equations introduced by
each method, assuming the implicit form of the dynamics is used. The only
exception is the PKT method, which, in accordance to \cite{Posa_ICRA2016}, is
written using the semi-implicit form. In all cases, we only give the variables
and equations introduced by the dynamics, so boundary or path constraints do not
appear in the table. By adding the number of variables minus the number of
equations in each method, it is easy to see that the solution space of the NLP
problem is of dimension $(N+1)\: n_u+d_\X$ in all transcriptions. Thus, for a
fixed $N$, all methods allow the same freedom to compute the optimal solution.

\begin{figure}[t!]
	\begin{center}
		\pstool[width=0.75\linewidth]{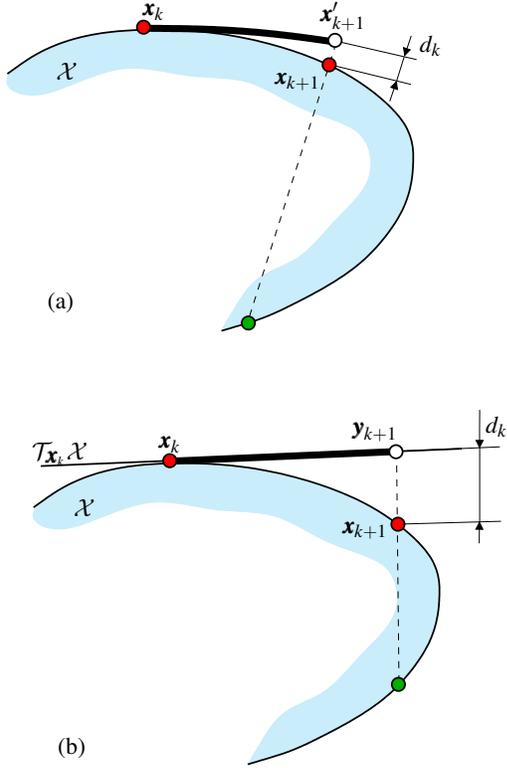}
		{
			\psfrag{d}[l]{\small $d_k$}	
			\psfrag{Txk}[l]{\small $\TX{\vr{x}_k}$}
			\psfrag{X}[l]{\small $\X$}
			\psfrag{xk}[l]{\small $\vr{x}_{k}$}
			\psfrag{xkpp}[l]{\small $\vr{x}_{k+1}'$}
			\psfrag{xkp}[l]{\small $\vr{x}_{k+1}$}
			\psfrag{yk}[l]{\small $ $} %
			\psfrag{yend}[l]{\small $\vr{y}_{k+1}$}		
			\psfrag{mu}[l]{\small $d$}		
			\psfrag{a}[l]{\small (a)}			
			\psfrag{b}[l]{\small (b)}		
		}
		\caption{Projection rays in the projection and local coordinates methods [(a)
			and (b), respectively]. Whereas in (a) the ray is orthogonal to $\X$, in (b)
			it is orthogonal to $\TX{\vr{x}_k}$. In both cases there may be more than one
			solution to the projection step to $\X$ (red and green points). The
			transcriptions should ensure the optimizer selects the red point for
			$\vr{x}_{k+1}$, i.e., the one that is closest to the projection point (in
			white). See the text for details. \label{fig:projectionIssue} }
	\end{center}
  \vspace{-0.25cm}
\end{figure}

\subsection{Ensuring proper projections}
\label{subsec:proper_projections}

For both the projection and local coordinates methods, we may find situations in
which a projection ray intercepts the~$\X$ manifold in multiple points (Fig.
\ref{fig:projectionIssue}). In these situations, we need to ensure that the
position of $\vr{x}_{k+1}$ chosen by the optimizer (red point) is the one that
is closest to the projection point (white point), otherwise the trajectory would
suddenly jump to farther regions in $\X$ (green point). To this end, we define
the distance $d_k$ from the projection point to $\vr{x}_{k+1}$ as
\begin{equation}
{d}_k = \lVert \vr{x}_{k+1}'-\vr{x}_{k+1} \rVert
\end{equation}
in the projection method, and as
\begin{equation}
{d}_k = \lVert (\vr{x}_{k}+\mt{U}_k(\vr{x}_k)\;\vr{y}_{k+1})-\vr{x}_{k+1} \rVert
\end{equation}
in the local coordinates method, and we add a small penalty term proportional to
\begin{equation}
	\label{eq:dk_penalty}
	\sum_{k=1}^{N} d_k^2
\end{equation}
in the cost function $C(\vr{w})$ in Eq. \eqref{eq:NLPcost}. In this way, the
optimizer selects the closest point to the projection point in the ray.

\subsection{Setting the boundary conditions}
\label{subsec:boundary}

Many trajectory optimization problems require fixing the end point of the
trajectory $\vr{x}_N$ to a particular state $\vr{x}_g$.  However, the direct
transcription of this constraint through imposing $\vr{x}_N=\vr{x}_g$ is not
suitable in constrained systems. In such systems, all transcription methods will
implicitly or explicitly constrain~$\vr{x}_N$ to be a point of $\X$. For
instance, in the basic method, only the drift prevents $\vr{x}_N$ to be on $\X$
once a control sequence is fixed, and other transcription methods explicitly
include the constraint $\vr{F}(\vr{x}_N)=\vr{0}$. In all cases, directly using
$\vr{x}_N=\vr{x}_g$ would result on an overconstrained system, thus violating
the required LICQ conditions~\cite{nocedal2006numerical}. A way around this
problem consists in replacing $\vr{x}_N=\vr{x}_g$ by

\begin{equation}
	\label{eq:projectedGoal}
	\mt{U}_g\tr(\vr{x}_N - \vr{x}_g) = \vr{0},
\end{equation}
where $\mt{U}_g$ is an $n_x \times d_\X$ matrix whose columns provide an
orthogonal basis of $\TX{\vr{x}_g}$. Equation \eqref{eq:projectedGoal}
constrains $\vr{x}_N$ to lie in the normal space of $\X$ at $\vr{x}_g$,  so it
removes the $d_\X$ degrees of freedom of $\vr{x}_N$ on $\X$. In this case,
however, a projection ambiguity may  appear which can be avoided including a
small penalty in the cost function proportional to $\|\vr{x}_N - \vr{x}_g\|^2$.
For the PKT, projection, and local coordinates methods, which include the
constraint  $\vr{F}(\vr{x}_N)=\vr{0}$, an alternative solution consists in
replacing this constraint by~$\vr{x}_N=\vr{x}_g$. Similiar considerations apply
when fixing $\vr{x}_0$ to a particular value.

\subsection{Finding an initial guess} \label{subsec:initial_guess}

Like most trajectory optimization methods, those we propose require reasonable
initial guesses of $\vr{u}(t)$ and $\vr{x}(t)$ to converge to a locally-optimal
solution. A typical way to find them consists in constructing an approximate
trajectory $\vr{x}(t)$ satisfying the boundary conditions (for example by
interpolating the start and goal states or using some educated guess), then
evaluating $\vr{x}(t)$ for $t = t_0, \ldots, t_N$ to obtain values for
$\vr{x}_0, \ldots, \vr{x}_N$, and finally deriving consistent actions $\vr{u}_0,
\ldots, \vr{u}_N$ by solving the inverse dynamics using Eq.~\eqref{eq:dyn3}.
This also provides initial guesses of $\vr{\lambda}_0, \ldots, \vr{\lambda}_N$,
which are needed if we resort to implicit forms of the dynamic equations (see
Section~\ref{subsec:implicit_vs_explicit}).

The previous initializations work well in simple situations, but in constrained
robots the joint angles tend to be highly coupled, so guessing their
trajectories is difficult, and they may be underactuated, or have singularities,
which prevents the solution of the inverse dynamics in
general~\cite{bordalba2021thesis}.  One way to circumvent these difficulties is
to resort to a randomized kinodynamic planner like the one in \cite{Bordalba_TRO2020,bordalba2021thesis},
which finds a dynamically-feasible trajectory respecting the force limits of the
actuators even across forward singularities. This trajectory will often be
jerky, and far from optimal, but in many cases it is good enough to allow the
convergence of the optimizer. This has been our method of choice in the examples
below (Section~\ref{sec:examples}). A good account of other initialization
techniques is given in \cite{Kelly_SIAM2017}.

\begin{figure}[t!]	
	\begin{center}
		\pstool[width=0.85\linewidth]{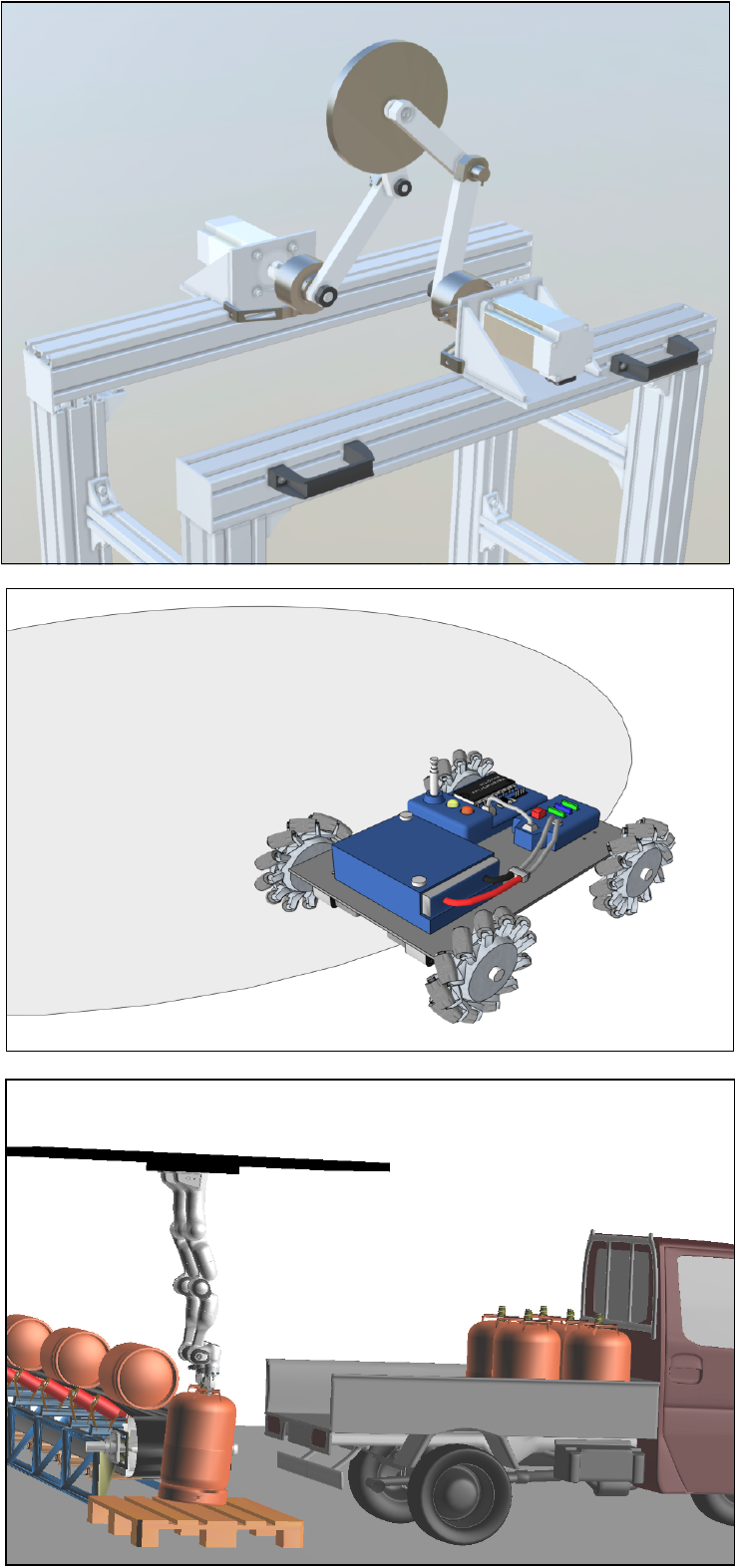}{}
		\caption{Examples. Top: a parallel robot used to lift a heavy load. 
			Middle: a robot with omni-directional wheels constrained to follow 
			a circular path. Bottom: two Franka Emika manipulators used to lift gas bottles onto a truck.}
		\label{fig:robots}
	\end{center}
  \vspace{-0.25cm}
\end{figure}

\begin{figure}[t!]
	\begin{center}
		\pstool[width=0.7\linewidth]{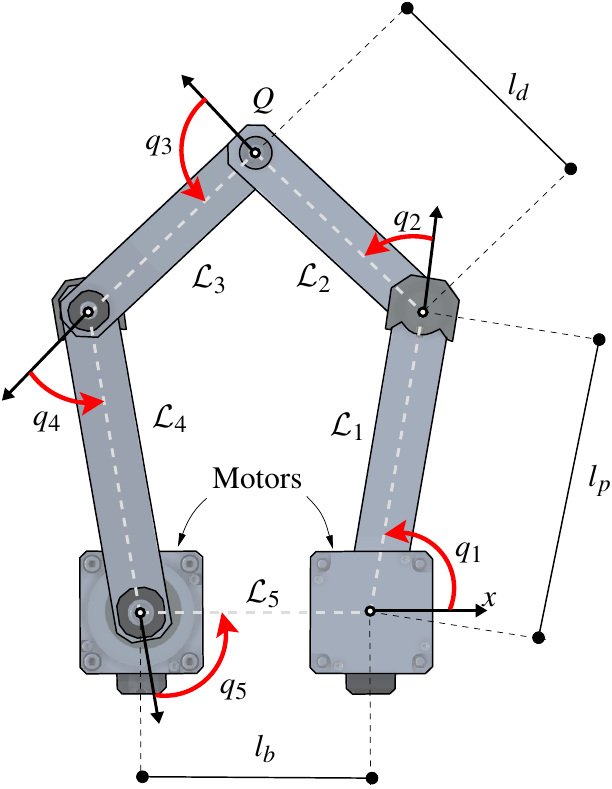}
		{
			\psfrag{Motors}[l]{\small Motors}	
			\psfrag{loop}[l]{\small loop}
			\psfrag{x}[l]{\small $x$}
			\psfrag{Q}[l]{\small $Q$}
			\psfrag{q1}[l]{\small $q_1$}
			\psfrag{q2}[l]{\small $q_2$}
			\psfrag{q3}[l]{\small $q_3$}
			\psfrag{q4}[l]{\small $q_4$}
			\psfrag{q5}[l]{\small $q_5$}
			\psfrag{lp}[l]{\small $l_p$}
			\psfrag{lb}[l]{\small $l_b$}
			\psfrag{ld}[l]{\small $l_d$}
			\psfrag{L1}[l]{\small $\Link{1}$}
			\psfrag{L2}[l]{\small $\Link{2}$}
			\psfrag{L3}[l]{\small $\Link{3}$}
			\psfrag{L4}[l]{\small $\Link{4}$}
			\psfrag{L5}[l]{\small $\Link{5}$}
		}
		
		\vspace{1mm}
		
		{
			\footnotesize
			\begin{tabular}{lcll}
				\toprule 
				Parameter							& Symbol &  Value	 	& Units						\\ 
				\toprule
				Base distance						& $l_b$	 &  $0.12$ 		& [m] 						\\ 
				Proximal link length 				& $l_p$  &  $0.20$ 		& [m] 						\\ 
				Distal link length 					& $l_d$  &  $0.15$ 		& [m] 						\\ 
				Mass of proximal link 				& $m_p$  &  $1.20$ 		& [kg] 						\\ 
				Mass of distal link 				& $m_d$  &  $0.90$ 		& [kg] 						\\ 
				Moment of inertia of proximal link 	& $I_p$  &  $0.007$ 	& [kg$\cdot$m$^2$] 			\\ 
				Moment of inertia of distal link 	& $I_d$  &  $0.002$ 	& [kg$\cdot$m$^2$] 			\\ 				
				Mass of the circular weight at $Q$	& $m_w$  &  $0.5$ 		& [kg] 						\\ 
				Viscous friction coefficient  		& $b$ 	 &  $0.07$ 		& [N$\cdot$m$\cdot$s$/$rad] \\
				Maximum torque of each motor 		& $\tau_{max}$ 	&  $1.4$  & [N$\cdot$m] 				\\
				\bottomrule
			\end{tabular}
		} \vspace{1mm} \caption{Geometry and dynamic parameters of the parallel robot. All joints are revolute. Joints
			$q_1$ and $q_5$ are actuated, and the remaining joints are passive. The motors are fixed to the ground, which acts as
			a fifth bar. All moments of inertia are given about the center of mass of the corresponding link. \label{fig:Scherbot} }
	\end{center}
\end{figure}

\subsection{Accuracy metrics} \label{subsec:solution_errors}

\begin{figure*}[t!]
\begin{center}
	\pstool[width=\textwidth]{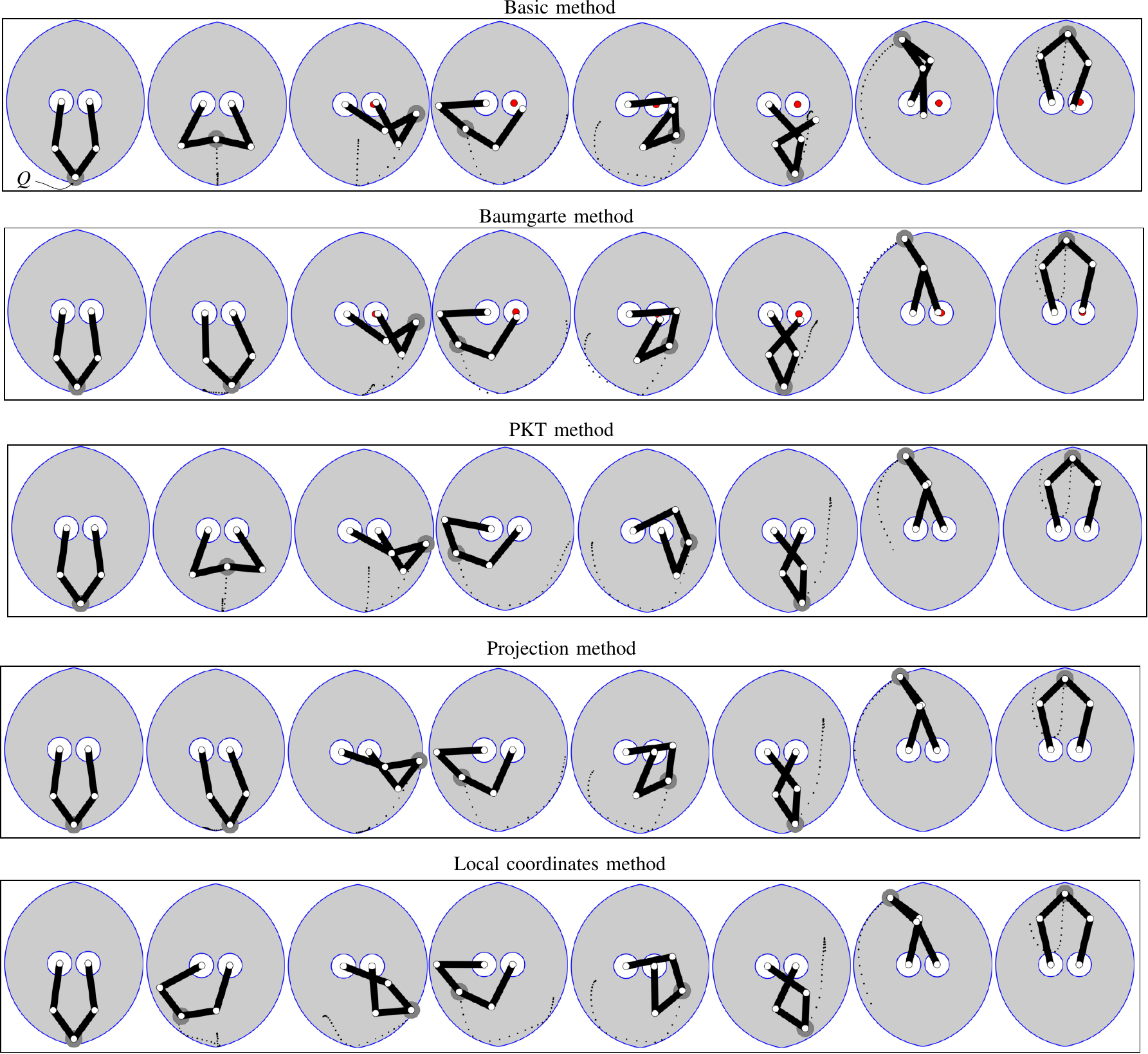}
	{
	  \psfrag{naive}[Bc]{\small Basic method} 
	  \psfrag{local}[Bc]{\small Local coordinates method} 
	  \psfrag{baumgarte}[Bc]{\small Baumgarte method} 
	  \psfrag{pkt}[Bc]{\small PKT method} 
	  \psfrag{projection}[Bc]{\small Projection method} 
	  \psfrag{Disassembled}[l]{\scriptsize Disassembled}
	  \psfrag{Slightly}[l]{\scriptsize Slightly}
	  \psfrag{disassembled}[l]{\scriptsize disassembled}		
	  \psfrag{Assembled}[l]{\scriptsize Assembled}
	  \psfrag{Q}[l]{\small $Q$}
  }
	\caption{Trajectories obtained by the five methods in the weight lifting
		task. See \cite{bordalba2022video} for a video
		of this figure. The grey area in each snapshot is the workspace of the robot,
		i.e., the set of positions that point $Q$ can attain. The two base joints are
		actuated. The trajectory from the basic method shows that, due to the
		accumulation of drift, the mechanism disassembles at the right motor joint (see
		the mismatch between the expected and obtained positions of this joint in
		several snapshots, in red and white respectively). In the trajectory from the
		Baumgarte method we see that, despite the stabilization of drift, the mechanism
		slightly disassembles and $Q$ leaves its workspace at times. In the PKT and
		projection methods, in contrast, the mechanism is kept almost assembled (only
		tiny disassemblies arise in between knot points in the video). The local
		coordinates method finally removes all disassemblies.  \label{fig:snapshots}
	}	
	\end{center}
\end{figure*}

To evaluate the quality of a trajectory it is essential to define proper
accuracy metrics. These metrics allow us to compare the five transcription
methods described in this paper. We here use two common error functions to
quantify how well $\vr{x}(t)$ and $\vr{u}(t)$ satisfy Eqs.~\eqref{eq:F}
and~\eqref{eq:explicitdynamics}. The logic is that if these two equations are
accurately fulfilled (both at the knot points and in between them) then the
spline for $\vr{x}(t)$ will provide an accurate representation of the motion
induced by $\vr{u}(t)$. Therefore, the lower the errors, the lower the control
effort needed to stabilize the trajectories a posteriori.

Specifically, we define the kinematic error as
\begin{equation}
e_K(t) = \lVert\vr{F}(\vr{x}(t)) \rVert,
\label{eq:kinematic_error}
\end{equation}
and the dynamic error as
\begin{equation}
e_D(t) = \lVert \vr{\dot{x}}(t) - \vr{f}(\vr{x}(t),\vr{u}(t)) \rVert.
\label{eq:dynamic_error}
\end{equation}
The averages of these two errors over $[0,t_f]$, 
\begin{eqnarray} \label{eq:int_kinematic_error}
E_K = \frac{1}{t_f}\int_{0}^{t_f} e_K(t) \; \textrm{d}t, \\ \label{eq:int_dynamic_error}
E_D = \frac{1}{t_f}\int_{0}^{t_f} e_D(t) \; \textrm{d}t,
\end{eqnarray}
provide global quantities summarizing the two types of errors over the whole
trajectories $\vr{x}(t)$ and $\vr{u}(t)$. In our implementation, these integrals
are computed using Shampine's adaptive quadrature method included in
Matlab~\cite{shampine2008vectorized}, under an absolute error tolerance of
$10^{-10}$.

\section{Examples} 
\label{sec:examples}

We next compare the performance of all methods in solving trajectory
optimization problems for the three systems shown in Fig.~\ref{fig:robots}: a
parallel robot used to lift a heavy weight, a Mecanum-wheeled vehicle following a
prescribed path, and a dual-arm system that loads gas bottles onto a
truck. Whereas the first system illustrates the methods' handling of holonomic
constraints (in this case a loop-closure constraint), the second shows how they
cope with nonholonomic constraints (the rolling contacts), and the third
provides a case with many degrees of freedom. All optimization problems have
been formulated as in Table~\ref{tab:counting}, and they have been solved using
a MacBook Pro with an Intel i9, 6-core processor running at 2.9 GHz. Except where
noted, the final time $t_f$ is fixed beforehand.

\begin{figure*}[p]
	\begin{center}
		\pstool[width=\textwidth]{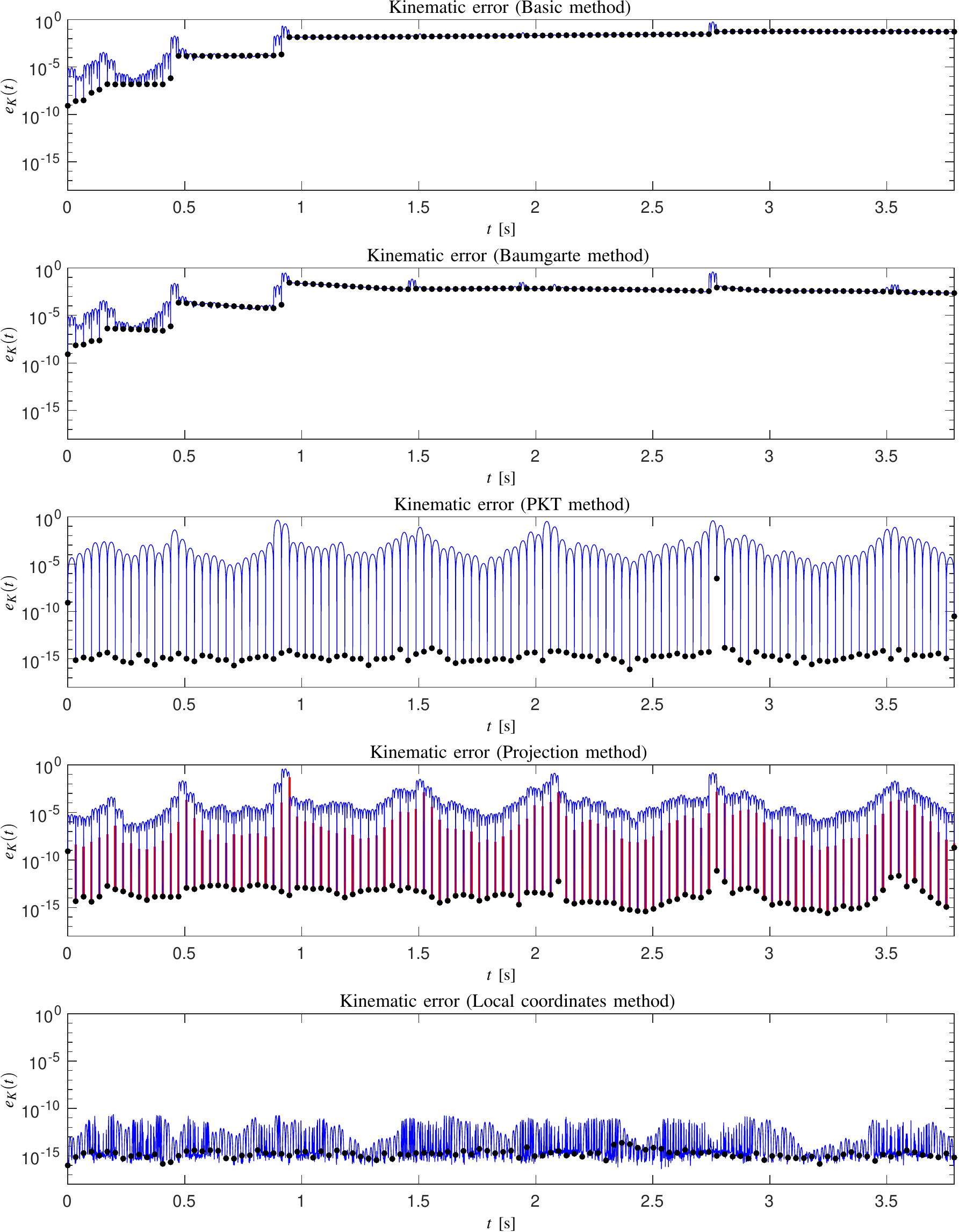} 
		{
			\psfrag{Knot points}[Bl]{\scriptsize Knot points}
			\psfrag{t [s]}[Bl]{\hspace{-2mm}\small $t$ [s]}
			\psfrag{ek}[Bl]{\small $e_K(t)$}
			\psfrag{dq}[Bl]{\small $\dq$ [rad/s]}		
			\psfrag{Kinematic error (Basic method)}[Bl]{\normalsize	  Kinematic error (Basic method)}
			\psfrag{Kinematic error (Baumgarte method)}[Bl]{\normalsize	  Kinematic error (Baumgarte method)}
			\psfrag{Kinematic error (PKT method)}[Bl]{\normalsize	  Kinematic error (PKT method)}
			\psfrag{Kinematic error (Projection method)}[Bl]{\normalsize	  Kinematic error (Projection method)}
			\psfrag{Kinematic error (Local coordinates method)}[Bl]{\normalsize	  Kinematic error (Local coordinates method)}
		}
		\caption{Logarithmic plot of the kinematic error $e_K(t)$ for the weight-lifting trajectories
			obtained by all methods (shown in blue and using $d=3$). The black dots indicate
			the values of $e_K(t)$ at the knot points. In the fourth plot, the red
			segments correspond to the projections from $\vr{x}_k'$ to $\vr{x}_k$ shown in
			Fig. \ref{fig:projection}. \label{fig:Scherbot_kinError}}
	\end{center}
	\vspace{2mm}
\end{figure*}

\begin{figure}[t!]
	\begin{center}
		\pstool[width=\linewidth]{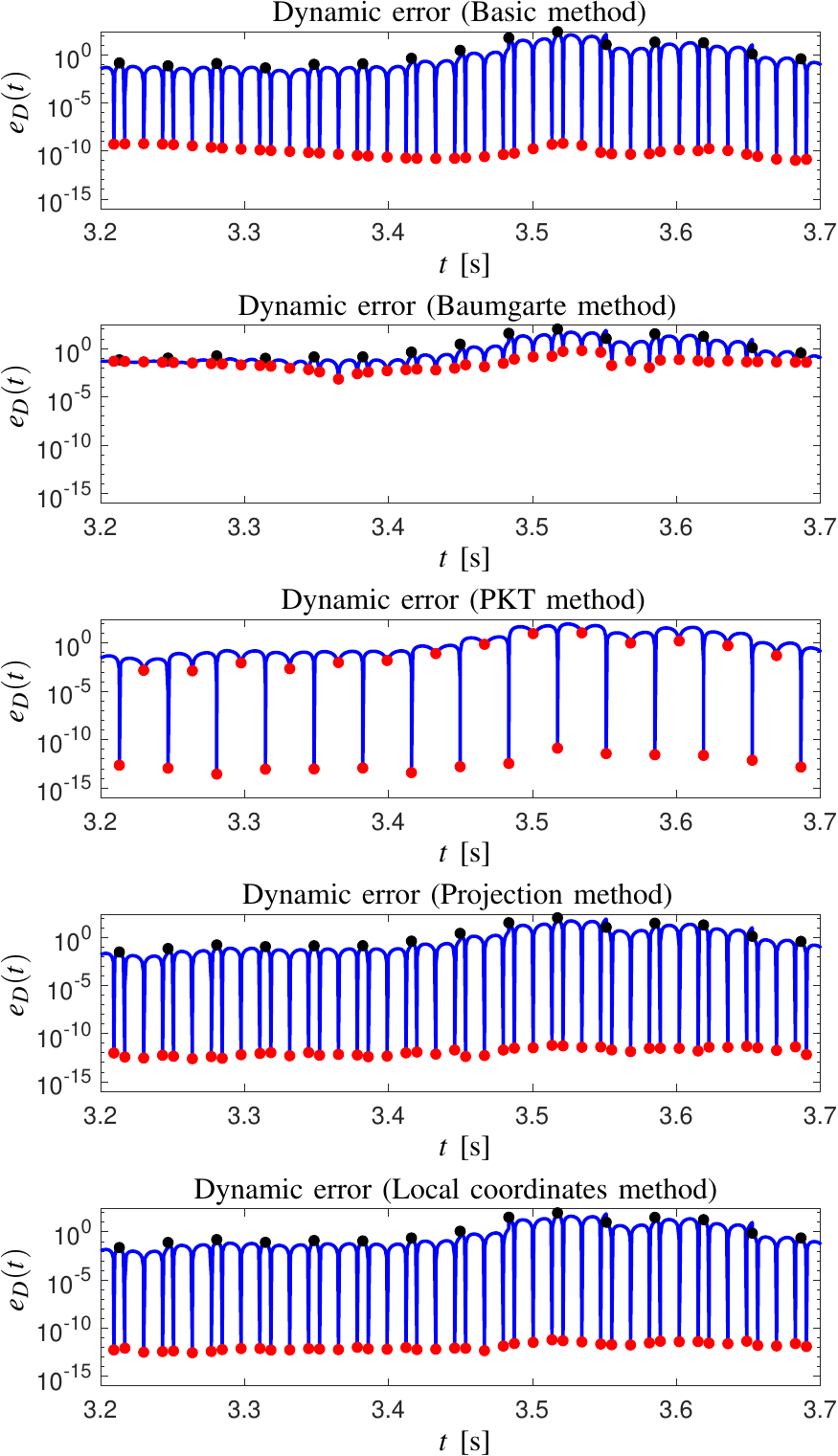} { \psfrag{Knot
		points}[Bl]{\tiny Knot points} \psfrag{Col. points}[Bl]{\tiny Col. points}
	\psfrag{t [s]}[Bl]{\small $t$ [s]} \psfrag{ed}[Bl]{\small $e_D (t)$}
	\psfrag{ek}[Bl]{\small $e_D (t)$} \psfrag{Dynamic error (Basic
		method)}[Bl]{\small Dynamic error (Basic method)} \psfrag{Dynamic error
		(Baumgarte method)}[Bl]{\small Dynamic error (Baumgarte method)}
	\psfrag{Dynamic error (PKT method)}[Bl]{\small Dynamic error (PKT method)}
	\psfrag{Dynamic error (Projection method)}[Bl]{\small Dynamic error
		(Projection method)} \psfrag{Dynamic error (Local coordinates method)}[Bl]{\small Dynamic
		error (Local coordinates method)} } \caption{Logarithmic plot of the dynamic
	error $e_D(t)$ for the weight-lifting task in all methods, using polynomial
	splines of degree $d=3$. We only depict $e_D(t)$ for the time span $[3.2,3.7]$
	to better appreciate this error at the collocation points (in red), but the
	trends are similar in the whole trajectory. The black dots indicate the value
	of $e_D(t)$ for the knot points $t_0, \ldots, t_N$. In the PKT method the knot
	points coincide with collocation points (those where $e_D(t)$ is negligible), so their black dots are occluded.
	\label{fig:Scherbot_dynError}}
  \vspace{-0.45cm}
	\end{center}
\end{figure}

\subsection{A system with holonomic constraints}

The parallel robot we consider has the geometry shown in Fig~\ref{fig:Scherbot}.
It consists of a closed loop of five links $\Link{1}, \ldots, \Link{5}$ pairwise
connected by revolute joints. In the figure, $q_i$ denotes the relative angle at
the $i$-th joint, which connects $\Link{i}$ with $\Link{i-1}$. Joints $1$ and
$5$ are actuated, allowing the control of the $(x,y)$ position of point $Q$, which
is regarded as the end effector. The rest of joints are passive. A heavy disk is
mounted on the axis at $Q$, as shown in Fig.~\ref{fig:robots}, top. The goal is
to lift this disk from a bottom position in which the robot is at rest, to an
upright position. To complicate the task, the robot is set to move on a vertical
plane, so it must overcome gravity, and we limit the motor torques to a small
range $[-\tau_\text{max},\tau_\text{max}]$ that impedes direct trajectories to
the goal. In all computations we assume the parameters given in
Fig~\ref{fig:Scherbot}.

In this robot, Eq.~\eqref{eq:phi} consists in the closure condition
imposed by the kinematic loop formed by the five links. This condition can be
expressed as
\begin{equation} 
	\label{eq:phi_Scherbot}
	\begin{bmatrix}
		l_p \; \textrm{c}(\bar{q}_1) + l_d\; \textrm{c}(\bar{q}_2) + 
			l_d \; \textrm{c}(\bar{q}_3)+l_p\; \textrm{c}(\bar{q}_4) + l_b \\
		l_p \; \textrm{s}(\bar{q}_1) + l_d\; \textrm{s}(\bar{q}_2) + 
			l_d \; \textrm{s}(\bar{q}_3)+l_p\; \textrm{s}(\bar{q}_4) \\
		s(\bar{q}_5) 
	\end{bmatrix}
	= 
	\begin{bmatrix} 
		0 \\ 0 \\ 0 
	\end{bmatrix},
\end{equation}
where $\textrm{s}(\cdot)$ and $\textrm{c}(\cdot)$ denote the sine and cosine of
their argument and $\bar{q}_i=\sum_{j=1}^i q_j$ is the angle of $i$-th link
relative to ground. In this case, Eq.~\eqref{eq:Bqdot} contains the time
derivative of Eq. \eqref{eq:phi_Scherbot} only, as all robot constraints are
holonomic. The generalized coordinates of this robot are $\vr{q} =
(q_1,\ldots,q_5)$, so $n_x = 10$, $n_p = n_v = 3$, and $\X$ is $4$-dimensional
in this problem. To set up Eq.~\eqref{eq:dyn4}, moreover, we have used the
methods in~\cite{featherstone2016dynamics}, which efficiently calculate all
terms intervening.

Fig.~\ref{fig:snapshots} shows weight-lifting trajectories obtained by all
methods using the running cost $L(\vr{x}(t),\vr{u}(t))=\vr{u}(t)\tr \vr{u}(t)$
for Eq. \eqref{eq:OCP_cost}. For a fair comparison, we have set $d=3$ in all
methods, as the PKT method can only work with cubic polynomials. As we see in
the figure, in all trajectories the torque limits force the generation of
swinging motions to reach the goal. The trajectory from the basic method shows
that the drift inevitably accumulates, so the mechanism disassembles and the
goal state cannot be reached (see the mismatch between the expected and obtained
positions of the right motor joint in several snapshots, in red and white
respectively). The trajectory from the Baumgarte method reveals that, although
the drift is mitigated, the mechanism still disassembles slightly. The PKT and
projection methods behave similarly, as both methods eliminate the drift at the
knot points, so only tiny disassemblies are visible in between such points (see
the the animation of this figure in~\cite{bordalba2022video}). The local
coordinates method finally eliminates any drift along the continuous time
trajectory.

The kinematic error corresponding to the previous trajectories is shown in
Fig.~\ref{fig:Scherbot_kinError}. Note that a large value of $e_K(t)$ implies
that Eq.~\eqref{eq:phi_Scherbot} is violated, which results in unrealistic
trajectories that are difficult to track with a controller. Thus, a method
resulting in low values of $e_K(t)$ is preferable. From the figure we see that
each method performs as expected. In the basic method, $e_K(t)$ tends to
accumulate over time, which in Fig.~\ref{fig:snapshots} corresponds to the
progressive disassembly of the right motor joint. In the Baumgarte method, the
stabilizing terms reduce the accumulation of drift from $t=1$ onwards, but many
knot points still deviate substantially from the manifold (black dots). In the
PKT and projection methods, instead, $e_K(t)\approx 0$  at the knot points, but
not in between them. Finally, $e_K(t)$ is negligible for all $t \in [0,t_f]$ in
the local coordinates method, so this is the most accurate of all methods.

To see how the different methods compare in terms of satisfying the collocation
constraints, Fig.~\ref{fig:Scherbot_dynError} provides the dynamic error
$e_D(t)$ for all methods in the time span $[3.2,3.7]$, which is representative
of the rest of the trajectory. As expected, in the basic  method the value of
$e_D(t)$ is negligible at the collocation points (red dots in the figure)
because it imposes the actual dynamics at such points. In the Baumgarte method,
in contrast, $e_D(t)$ is much larger at the collocation points, because this
method enforces a modified version of the dynamics to mitigate the problem of
drift. The PKT method is an improvement over the Baumgarte method because
$e_D(t)$ is negligible in one of every two collocation points. In the projection
and local coordinates methods, instead, $e_D(t)$ is negligible in all
collocation points.

\begin{table*}[t!]
	\caption{Performance statistics for the weight lifting task in the parallel robot as $d$ increases}  \label{tab:scherbot}
	\centering
	\setlength{\tabcolsep}{9pt}
	\begin{tabular}{lrrrrrrrrr} 
			\toprule
			Method & $d$ & $n_w$ & $n_c$ & $E_K \quad\quad$ & $E_D\;\;$ & $C^*\;$ & $t_\textrm{opt}$ [s] & $n_i$ & $t_i$ [s]\\
			\midrule
			Basic    & $2$ &  5388 &  5166 & $31.44\cdot 10^{-3}$ & $58.57$ & $0.73$ & $3.02$ & $55$ & $0.055$ \\
			& $3$ &  7404 &  7182 & $29.06\cdot 10^{-3}$ & $43.70$ & $1.07$ & $5.65$ & $54$ & $0.105$ \\
			& $4$ & 9420 & 9198 &  $8.40\cdot 10^{-3}$ & $26.46$ & $1.74$ & $15.83$ & $76$ & $0.208$ \\
			\midrule
			Baumgarte & $2$ & 5388 &  5166 &$19.09\cdot 10^{-3}$ & $61.21$ & $0.64$ & $4.14$ & $56$ & $0.074$ \\
         	& $3$ &  7404 &  7182 & $15.47\cdot 10^{-3}$ & $31.59$ & $1.68$ & $9.75$ & $71$ & $0.137$ \\
   	       	& $4$ & 9420 & 9198 &   $2.38\cdot 10^{-3}$ & $14.12$ & $2.06$ & $18.53$ & $98$ & $0.189$ \\   	       	
			\midrule
			PKT      &  3   &  2367 &  2145 & $10.90\cdot 10^{-3}$ & $33.19$ & $0.61$ & $6.69$ & $46$ & $0.145$ \\
			\midrule
      		Projection & $2$ &  7180 &  6958 & $11.47\cdot 10^{-3}$ & $63.90$ & $0.61$ & $4.02$ & $54$ & $0.074$ \\
             & $3$ & 9196 & 8974 & $4.10\cdot 10^{-3}$ & $30.91$ & $1.77$ & $4.89$ & $47$ & $0.104$ \\
             & $4$ & 11212 & 10990 & $1.37\cdot 10^{-3}$ & $14.30$ & $2.05$ & $13.54$ & $72$ & $0.188$ \\
             \midrule
      		Local coordinates    & $2$ &  6284 &  6062 & $4.13\cdot 10^{-13}$ & $39.36$ & $2.05$ & $20.51$ & $80$ & $0.256$ \\
             & $3$ & 10698 & 10426 & $5.04\cdot 10^{-13}$ & $18.36$ & $2.07$ & $16.14$ & $65$ & $0.248$ \\
             & $4$ & 11212 & 10990 & $3.82\cdot 10^{-13}$ & $10.03$ & $2.09$ & $21.80$ & $78$ & $0.279$ \\			
             \bottomrule
	\end{tabular}
  \vspace{-0.2cm}
\end{table*}

\begin{figure}[t!]
	\begin{center}
		\pstool[width=1\linewidth]{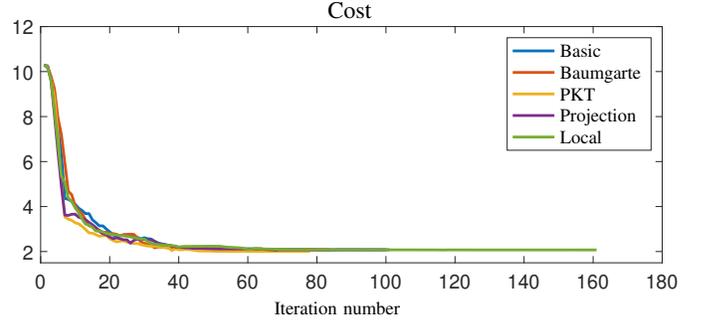} 
		{
			\psfrag{Iteration number}[Bl]{\scriptsize Iteration number}
			\psfrag{Basic}[Bl]{\scriptsize Basic}
			\psfrag{Baumgarte}[Bl]{\scriptsize Baumgarte}
			\psfrag{Local}[Bl]{\scriptsize Local}			
			\psfrag{Projection}[Bl]{\scriptsize Projection}
			\psfrag{PKT}[Bl]{\scriptsize PKT}
			\psfrag{Cost}[Bl]{\small Cost}
			\psfrag{Primal feasibility}[Bl]{\small Primal feasibility}
			\psfrag{Dual feasibility}[Bl]{\small Dual feasibility}
		}	
		\caption{Evolution of the cost as a function of the
			     iteration number.} \label{fig:convergence}
	\end{center}
  \vspace{-0.4cm}
\end{figure}

Table \ref{tab:scherbot} provides global performance measures for the five
methods when using polynomials of increasing degree~$d$ (except for the PKT
method, which requires $d=3$) using $N=112$ and $\Delta t = 0.034$~[s] in all
cases. For each value of~$d$ we provide the number of variables $n_w$ and
constraints~$n_c$ in the transcriptions, the average integral kinematic and
dynamic errors, $E_K$ and $E_D$, given by Eqs.~\eqref{eq:int_kinematic_error}
and \eqref{eq:int_dynamic_error} respectively, the cost $C^*$ of the optimized
trajectory (as a reference, the initial guess has $C=10.54$), the CPU time
$t_\textrm{opt}$ used to solve the NLP problem, the number of optimization
iterations $n_i$, and the average time per iteration $t_i$. As expected,
both~$E_K$ and~$E_D$ decrease when increasing $d$ in each method. For a same
value of $d$, the projection method (and also the PKT method when $d=3$) has
values of $E_K$ that are smaller than those of the basic and Baumgarte methods.
In contrast, $E_K$ is negligible in the local coordinates method, and only
depends on the tolerance used by the optimizer when solving
Eq.~\eqref{eq:psiImpl}. Also for a same $d$, the local coordinates method has
the smallest value of~$E_D$, which is in agreement with its higher accuracy.
This higher accuracy comes at the cost of a larger optimization time due to
computing the tangent space bases for all knot points in each iteration (see
Appendix~\ref{app:tangent_basis}). As a result, the cost of this operation
dominates the execution time for this method. This is why its time per iteration
is almost constant in the tests, irrespective of~$d$. This is not the case for
the rest of methods. Note that $n_e$ and $n_w$ indicate the size of the
corresponding NLP problem. The projection and the local coordinates methods
define larger NLP problems, but this does not necessarily translate into larger
execution times. This effect is remarkable for the PKT method, which is the one
with smallest programs, but not the one with the lowest execution times. This
can be attributed to the fact that the PKT method uses the semi-implicit form of
the dynamics, instead of the fully-implicit one
(Section~\ref{subsec:implicit_vs_explicit})

As shown in the companion video of this paper (see \cite{bordalba2022video}),
the different methods obtain solution trajectories in a same homotopy class. The
slight differences between the trajectories, and in the values of~$C^*$, can be
attributed to the violation of the kinematic constraints, as only the local
coordinates method can find a solution that is fully compliant with such
constraints. As the precision is increased, either by increasing $d$ or by
decreasing $\Delta t$, the kinematic error decreases and all methods converge to
a same value $C^*$. This is illustrated in Fig.~\ref{fig:convergence} which
shows the evolution of the cost for $d=3$, but using a smaller time step $\Delta
t=0.01$~[s]. Clearly, all methods converge to a solution with the same cost.

\begin{figure}[t!]
	\begin{center}
		\pstool[width=1\linewidth]{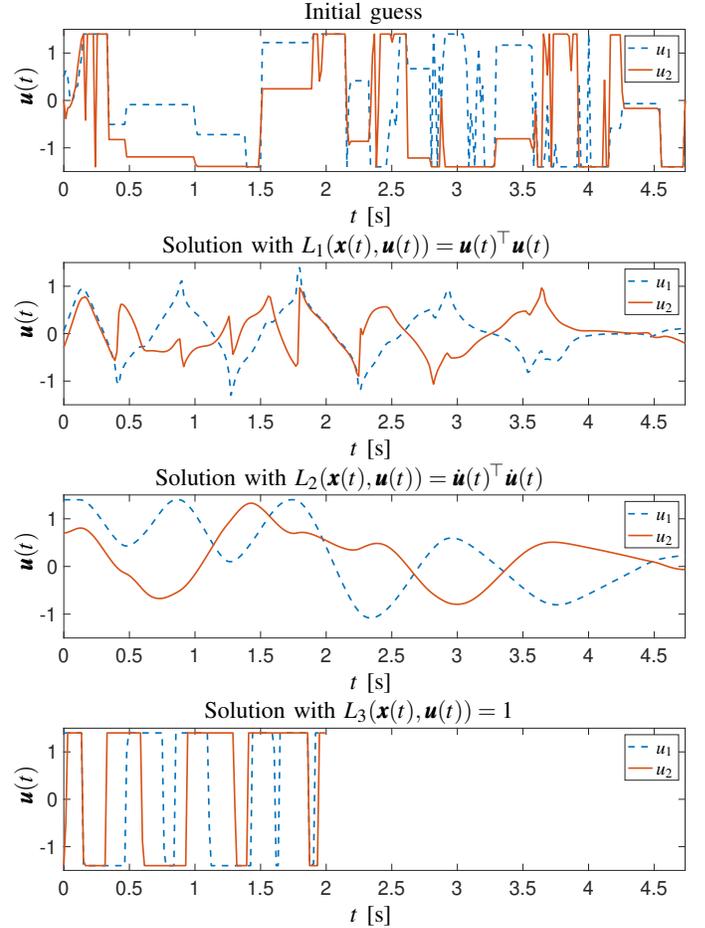} 
		{
			\psfrag{t}[Bc]{\small $t$ [s]}
			\psfrag{u}[Bl]{\small $\vr{u}(t)$}	
			\psfrag{u1}[Bl]{\scriptsize ${u}_1$}	
			\psfrag{u2}[Bl]{\scriptsize ${u}_2$}	
			\psfrag{ini}[Bc]{\small  Initial guess}
			\psfrag{m du}[Bc]{\small  Solution with $L_2(\vr{x}(t),\vr{u}(t))=\vr{\dot{u}}(t)\tr\vr{\dot{u}}(t)$}
			\psfrag{m u}[Bc]{\small   Solution with $L_1(\vr{x}(t),\vr{u}(t))=\vr{u}(t)\tr\vr{u}(t)$}
			\psfrag{m t}[Bc]{\small  Solution with $L_3(\vr{x}(t),\vr{u}(t))=1$}
		}	
		\caption{Top plot: the initial guess of the action trajectory $\vr{u}(t)$. 
			Bottom plots: the optimized function $\vr{u}(t)$ obtained by the projection method with $d=4$ when using
			different running costs.
			See \cite{bordalba2022video}
			for the animated trajectories corresponding to each cost.
			\label{fig:Scherbot_input}}
	\end{center}
  \vspace{-0.3cm} 
\end{figure}

\begin{table*}[t!]
	\caption{Performance statistics for the circle following task in the wheeled robot when $d$ increases} \label{tab:omnibot} 
		\centering	
		\begin{tabular}{lrrrrrrrrr} 
			\toprule
			Method       & $d$ & $n_w$ & $n_c$ &   $E_K \quad\quad$    & $E_D\;\;$ & $C^*\;$ & $t_\textrm{opt}$ [s] & $n_i$ & $t_i$ [ms] \\
			\midrule		
			{Basic} & 2 & 630 & 600 & $1384.6\cdot 10^{-3}$ & 97.4 & 2188.3 & 0.1 & 30 & 4.1 \\
			& 3 & 855 & 825 & $358.3\cdot 10^{-3}$ & 18.9 & 2191.6 & 0.2 & 33 & 4.8 \\
			& 4 & 1080 & 1050 & $62.5\cdot 10^{-3}$ & 3.5 & 2191.7 & 0.1 & 27 & 5.4 \\
			\midrule
			{Baumgarte} & 2  & 630 & 600 & $1366.4\cdot 10^{-3}$ & 98.0 & 2188.3 & 0.1 & 27 & 2.8 \\
			& 3 &  855 & 825 & $344.8\cdot 10^{-3}$ & 18.7 & 2191.6 & 0.2 & 28 & 6.6 \\
			& 4 & 1080 & 1050 & $61.1\cdot 10^{-3}$ & 3.4 & 2191.7 & 0.2 & 29 & 5.9 \\
			\midrule
			{Projection} & 2 & 792 & 762 & $1094.1\cdot 10^{-3}$ & 83.1 & 2189.0 & 0.2 & 39 & 5.1 \\
			& 3 & 1017 & 987 & $335.7\cdot 10^{-3}$ & 20.0 & 2191.0 & 0.4 & 54 & 6.6 \\
			& 4 & 1242 & 1212 & $55.8\cdot 10^{-3}$ & 3.4 & 2191.7 & 0.4 & 48 & 8.9 \\
			\midrule
			{Local coordinates} & 2 & 810 & 780 & $1.7\cdot 10^{-13}$ & 46.4 & 2193.4 & 1.2 & 23 & 50.4 \\
			& 3 & 1125 & 1095 & $1.7\cdot 10^{-13}$ & 3.9 & 2195.1 & 3.9 & 64 & 61.3 \\
			& 4& 1440 & 1410& $4.0\cdot 10^{-13}$ & 0.2 & 2195.1 & 2.7 & 29 & 92.2 \\
			\bottomrule		
		\end{tabular}
\end{table*}

Fig. \ref{fig:Scherbot_input} shows that the methods can cope with
different cost functions, including those yielding bang bang controls. The
figure shows the initial guess of the action trajectory (top plot) and the
optimized action trajectories~$\vr{u}^*(t)$ that we obtain when using the
following running costs in Eq.~\eqref{eq:OCP_cost}:
\begin{align}
L_1(\vr{x}(t),\vr{u}(t)) &= \vr{u}(t)\tr\vr{u}(t), \\
L_2(\vr{x}(t),\vr{u}(t)) &= \dot{\vr{u}}(t)\tr\dot{\vr{u}}(t), \\
L_3(\vr{x}(t),\vr{u}(t)) &= 1.
\end{align}
As we see, the use of ${L}_1$ results in an action trajectory $\vr{u}^*(t)$ that
is smoother in comparison to the initial guess, so the control signal will be
easier to follow. If further smoothing is needed, we can use $L_2$, which will
minimize the derivative of $\vr{u}(t)$. The curves in the third plot of
Fig.~\ref{fig:Scherbot_input} are clearly smoother than those in the second
plot. Finally, if we need to minimize the total trajectory time, we can free
$t_f$ and use ${L}_3$ to obtain the optimized trajectory $\vr{u}^*(t)$ of the
fourth plot. A bang-bang control arises in which at least one motor
works either at its upper or lower torque limit, and the robot achieves the goal
state in only two seconds. The plots were obtained with the projection method,
but similar results are achieved with the local coordinates method.

\subsection{A system with nonholonomic constraints} \label{sec:omni}

\begin{figure}[t!]		
	\begin{center}
		\pstool[width=0.7\linewidth]{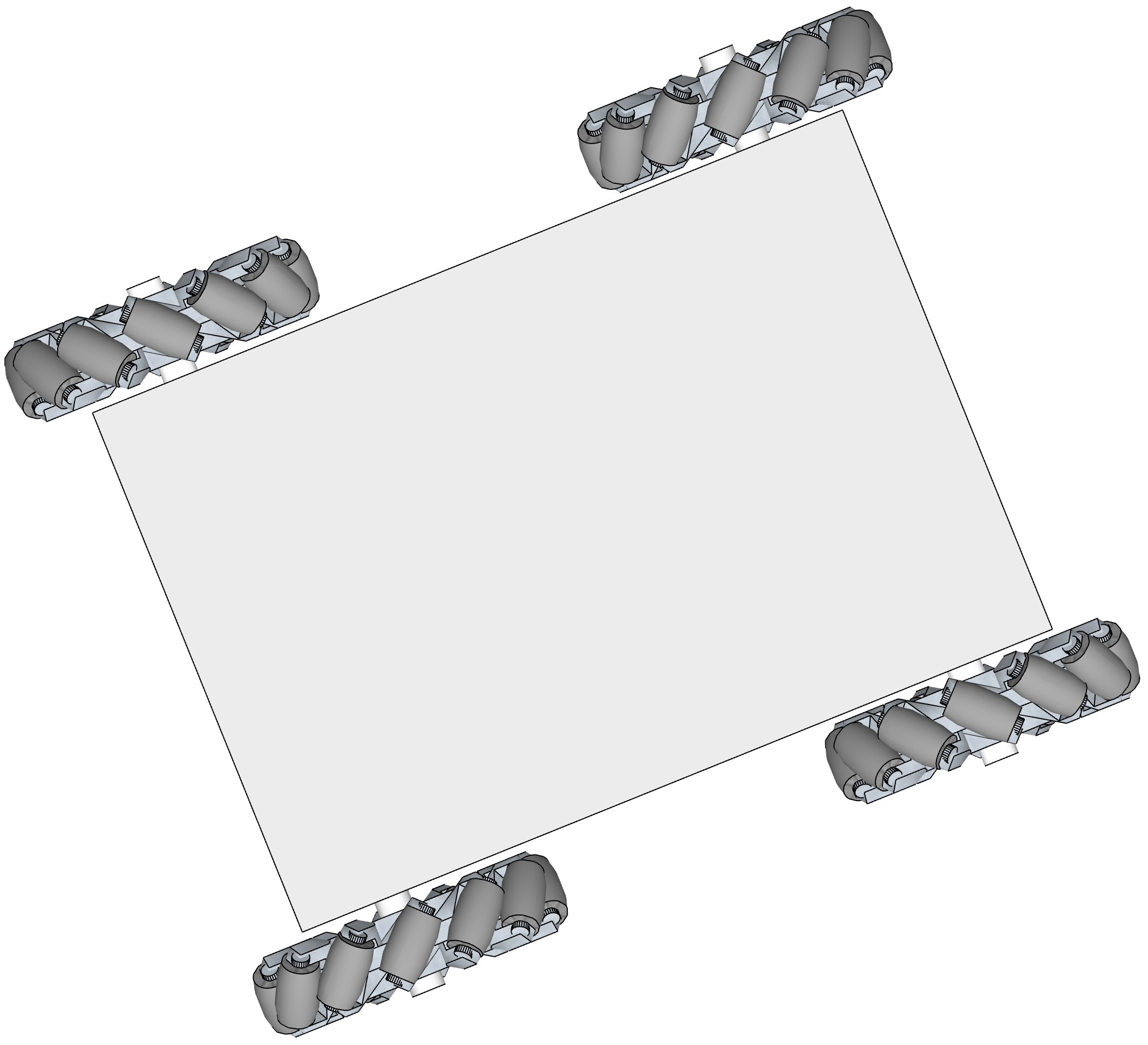}
		{
			\psfrag{fid1}[l]{\small $\dot{\varphi}_1$}
			\psfrag{fid2}[l]{\small $\dot{\varphi}_2$}
			\psfrag{fid3}[l]{\small $\dot{\varphi}_3$}
			\psfrag{fid4}[l]{\small $\dot{\varphi}_4$}
			\psfrag{l}[l]{\small $l$}
			\psfrag{L}[l]{\small $L$}
			\psfrag{r}[l]{\small $r$}				
			\psfrag{G}[l]{\small $G=(x,y)$}	
		}
		
		\vspace{5mm}
		
		{
		\footnotesize
		\begin{tabular}{lcll}
			\toprule 
			Parameter							& Symbol &  Value	 	& Units						\\ 
			\toprule
			Side length of the chassis			& $l$ 	 &  $0.21$ 		&[m] 						\\ 
			Wheel radius 						& $r$ 	 &  $0.066$ 	&[m] 						\\ 
			Vehicle mass 						& $m$ 	 &  $15.75$ 	&[kg] 						\\ 
			Moment of inertia of the chassis 	& $I_z$  &  $0.461$ 	&[kg$\cdot$m$^2$] 			\\ 
			Moment of inertia of the wheels 	& $I_w$  &  $0.0026$ 	&[kg$\cdot$m$^2$] 			\\ 
			Viscous friction at the wheel axes 	& $b$ 	 &  $0.1$ 		&[N$\cdot$m$\cdot$s$/$rad] 	\\
			\bottomrule
		\end{tabular}
		\vspace{2mm}
		}
		
		\caption{Geometric and dynamic parameters of the Mecanum-wheeled robot in the second example. They correspond to a Mecanum  Wheel  Vectoring  Robot  IG42  from  SuperDroid Robots Inc. See~\cite{moreno2021model} for details. The moments of inertia $I_z$ and $I_w$ are relative to an axis perpendicular to the chassis and wheel planes, meeting $G$ and the wheel centers respectively. \label{fig:omnibot-geom} }
	\end{center}
\end{figure}

To illustrate the proposed methods on nonholonomic systems, we next apply them
to a Mecanum-wheeled robot with the parameters shown in
Fig~\ref{fig:omnibot-geom}. The vehicle pose is given by the $(x,y)$ coordinates
of the centre of mass $G$ of the vehicle, and by the angle $\psi$ of the
chassis, all relative to an absolute frame. We denote the wheel angles as
$\varphi_1, \ldots,\varphi_4$.

In this example, the robot's position is constrained to a circle of
radius $R$ centered at the origin, so the trajectory must fulfil
\begin{equation}
x(t)^2+y(t)^2=R^2.
\label{eq:circle_cont}
\end{equation}
The chassis orientation $\psi(t)$ can freely be chosen instead, as the vehicle
has omni-directional wheels. In all tests we use $R = 2.5$ [m].

Note that Eq.~\eqref{eq:circle_cont} cannot be considered as part of
Eq.~\eqref{eq:phi} in this case because it is not intrinsic to the robot
structure. Instead, it must be viewed as a path constraint modelled by
Eq.~\eqref{eq:NLPpath}. In the transcribed problem, thus,
Eq.~\eqref{eq:circle_cont} will intervene in the form
\begin{equation}
	\label{eq:circle}
	x_k^2+y_k^2=R^2, \hspace{6mm} k=1,\ldots,N.
\end{equation}

To formulate the kinematic constraints, we denote the chassis configuration by
$\vr{q}_r=(x,y,\psi)$ and the wheels configuration by
$\vr{q}_w=(\varphi_1,\ldots,\varphi_4)$. The robot configuration is then given
by $\vr{q}=(\vr{q}_r,\vr{q}_w)$. As is common in mobile robots, we wish the
wheels do not slip, so according to \cite{moreno2021model} $\vr{q}_r$ and
$\vr{q}_w$ must be coupled by the nonholonomic constraint
\begin{equation}
	\mt{K}\; \mt{R}_Z\tr(\psi)  \; \dq_r - \dq_w = \vr{0},
	\label{eq:rolling_contact}
\end{equation}
where
\begin{equation*}
	\mt{K} = \frac{1}{r} 
	\begin{bmatrix}
		1 & -1 & -2 \: l        \\
		1 & \ \ 1 & \ \ 2 \: l  \\
		1 & \ \ 1 & -2 \: l     \\
		1 & -1 & \ \ 2 \: l 
	\end{bmatrix}
\end{equation*}
and $\mt{R}_Z(\psi)$ is the $3\times 3$ rotation of angle $\psi$ about the~$Z$
axis.

\begin{table*}[t!]
	\caption{Performance statistics for the truck loading task in the dual-arm system when $d$ increases} \label{tab:pandas}
		\centering	
		\begin{tabular}{lrrrrrrrrr} 
			\toprule
			Method     & $d$ & $n_w$ & $n_c$ & $E_K \quad\quad$ & $E_D\;\;$ & $C^*\;$ & $t_\textrm{opt}$ [s] & $n_i$ & $t_i$ [s]\\
			\midrule
			Basic
			& $2$ & 5838 & 5252 & $1.17\cdot 10^{-3}$ & $0.63$ & $11699.49$ & $392.70$ & $83$ & $4.731$ \\
			& $3$ & 7854 &	7268 & $0.17\cdot 10^{-3}$ & $0.13$ & $11731.70$ & $1221.90$ & $100$ & $12.219$ \\
			& $4$ & 9870 &	9284 & $0.12\cdot 10^{-3}$ & $0.11$ & $12418.45$ & $1815.59$ & $96$ & $18.912$ \\
			\midrule
			Baumgarte
			& $2$ & 5838 & 5252 & $0.86\cdot 10^{-3}$ & $0.63$ & $11700.70$ & $833.62$ & $126$ & $6.616$ \\
			& $3$ & 7854 &	7268 & $0.15\cdot 10^{-3}$ & $0.13$ & $11731.65$ & $1264.76$ & $181$ & $6.988$ \\
			& $4$ & 9870 &	9284 & $0.09\cdot 10^{-3}$ & $0.11$ & $12418.51$ & $1090.45$ & $81$ & $13.462$ \\
			\midrule
			PKT
			& $3$ & 2568 & 1982 & $3.24\cdot 10^{-3}$ & $1.89$ & $13058.05$ & $792.99$ & $611$ & $1.298$ \\
			\midrule
			Projection
			& $2$ & 7518 & 6932 & $1.16\cdot 10^{-3}$ & $1.61$ & $12425.24$ & $476.90$ & $700$ & $0.681$ \\
			& $3$ & 9534 &	8948 &  $0.14\cdot 10^{-3}$ & $0.13$ & $11731.23$ & $492.29$ & $453$ & $1.087$ \\
			& $4$ & 11550 &	10964 & $0.03\cdot 10^{-3}$ & $0.04$ & $11729.36$ & $1546.32$ & $457$ & $3.384$ \\			
			\midrule
			Local coordinates      
			& $2$ &  7182 &	6596 & $0.10\cdot 10^{-13}$ & $0.69$ & $12039.85$ & $28.84$ & $63$ & $0.458$ \\
			& $3$ &  9870 &	9284 &$0.12\cdot 10^{-13}$ & $0.15$ & $12979.68$ & $52.56$ & $81$ & $0.649$ \\
			& $4$ & 12558 &	11972 &$0.13\cdot 10^{-13}$ & $0.04$ & $12603.92$ & $56.18$ & $64$ & $0.878$ \\
			\bottomrule
		\end{tabular}
\end{table*}
	
\begin{figure*}[t!]
		\begin{center}
			\pstool[width=\textwidth]{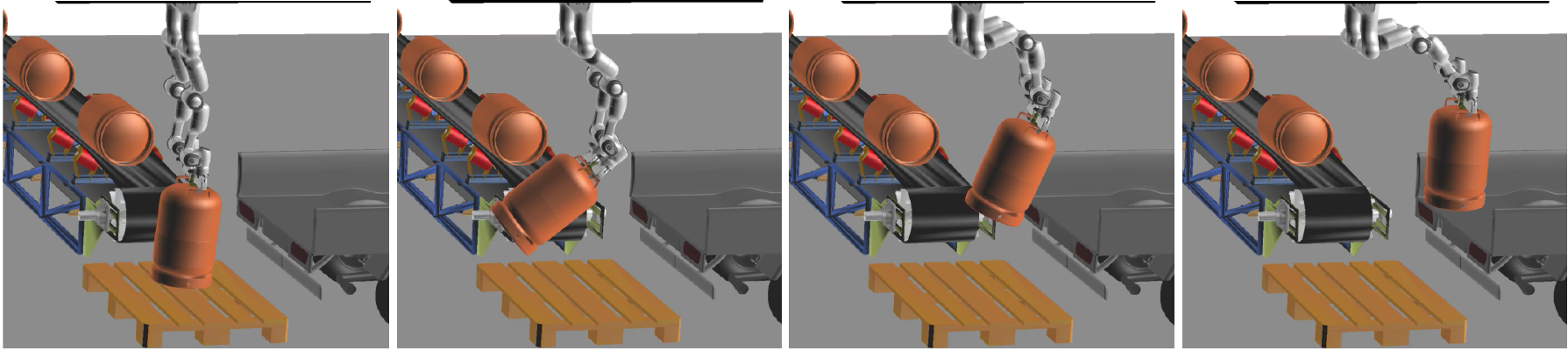}{}
			\caption{A trajectory obtained for the truck loading task in the dual arm system using the local 
				coordinates method and $d=3$. To gain momentum, the robot first moves the bottle backwards, 
				to then lift it forward onto the truck. The companion video of this paper provides an animated 
				version \cite{bordalba2022video}. \label{fig:snapshots_panda}}
		\end{center}
\end{figure*}

The optimizer must guarantee that Eq.~\eqref{eq:rolling_contact} is fulfilled as
closely as possible in the final trajectory, otherwise slippage of the
robot relative to the ground occurs, which generates odometry problems during
navigation. This equation can be brought into the form of Eq.~\eqref{eq:Bqdot}
by defining
\begin{equation}
	\mt{B}(\vr{q})=[\mt{K}\mt{R}_Z\tr(\psi) \quad -\mt{I}_4],
\end{equation}
where $\mt{I}_4$ is the $4\times4$ identity matrix. By taking the time
derivative of Eq.~\eqref{eq:rolling_contact} we obtain Eq.~\eqref{eq:dotdotphi}
in the form
\begin{equation}\label{eq:dotdotphi_omni}
	\mt{B}(\vr{q}) \; \ddq = -
	\mt{K}\mt{\dot{R}}_Z\tr(\psi) \; \dq_r.
\end{equation}
Using the derivations in~\cite{moreno2021model}, Lagrange's equation for
this robot is then given by
\begin{equation} 
	\label{eq:dynamics_omni}
	\mt{M}\ddq+\mt{B}(\vr{q})\tr \vr{\lambda} = \vr{\tau}(\vr{u},\dq),
\end{equation}
where 
\begin{equation*}
	\mt{M} =
	\begin{bmatrix} 
		\mt{M}_r &  \\
		& \mt{M}_w
	\end{bmatrix},
	\quad
	\mt{M}_r =
	\begin{bmatrix} 
		m & &  \\
		& m & \\
		& & I_z
	\end{bmatrix},
	\quad
	\mt{M}_w = I_w \Id{4},
\end{equation*}
and
\begin{equation}
	\vr{\tau}(\vr{u},\dq) = 
	\begin{bmatrix}
		\vr{0} \\
		\vr{u}-b\dq_w
	\end{bmatrix}. 
	\label{eq:bqw}
\end{equation}
In these equations, $\vr{u}=(u_1 , u_2 , u_3 ,u_4)$ is the vector of wheel
torques applied by the motors, and $b\dq_w$ models the viscous friction at the
wheels. Thus, in this example $n_u=4$, \mbox{$n_x=14$}, $n_p=0$, $n_v=4$, and
$d_\X=10$. As explained in Section \ref{sec:formulation},
Eqs.~\eqref{eq:dotdotphi_omni} and~\eqref{eq:dynamics_omni} can be combined to
find the explicit form of the dynamics equation
\mbox{$\vr{\dot{x}}=\vr{f}(\vr{x},\vr{u})$}.

Table~\ref{tab:omnibot} provides the performance statistics for this test case
using $N=9$, \mbox{$\Delta t = 0.2$~[s]}, and the running cost
\begin{equation}
	L(\vr{x}(t),\vr{u}(t))=\vr{u}(t)\tr\vr{u}(t),
\end{equation}
under increasing values of $d$. Results from the PKT method do not appear in
this table because, as we said before, it cannot be applied to nonholonomic
systems. All methods converge to trajectories of a similar cost in this example,
and the results confirm that the projection and local coordinates methods reduce
the kinematic error, or make it negligible, respectively. While the latter
method is again more costly, it also achieves a remarkable
decrease in the dynamic error. Compare the values of $E_D$  for $d=3$ or $4$ for
example.

\subsection{A system with many degrees of freedom} \label{sec:pandas}

To compare the methods in a system with many degrees of freedom, we now address
a task in which two Franka Emika 7-DOF arms have to move a gas bottle
cooperatively. Initially, the two arms are holding the bottle in a bottom
position at rest, and the goal is to load it onto a truck, arriving with zero
velocity at the goal configuration. The grippers in each arm rigidly
grasp the bottle, so only the robot joints can move.

For this example, the $\vr{q}$ vector is formed by the $14$ internal joint
angles, and Eq.~\eqref{eq:phi} contains $6$ loop-closure constraints. Thus, $n_x
= 28$, $n_p = n_v = 6$, and $d_\X = 16$. The mass of the bottle is twice the
added payload of the two arms, so by designing this trajectory, we allow the
system to move much beyond its static capabilities. 

Table~\ref{tab:pandas} provides performance statistics for the solutions we
obtain using $N=42$, \mbox{$\Delta t = 0.04$~[s]}, and the running cost
$L(\vr{x}(t),\vr{u}(t))=\vr{\dot{u}}^\top \vr{\dot{u}}$ under increasing values
of $d$. We use this running cost to promote trajectories with smooth
accelerations, which are useful to reduce stress, wear and tear of the robot
arms. With the aim of  obtaining highly accurate results without incurring in a
too high computational cost, in this example we use the output trajectory from
the projection method as the input for the local coordinates method, obtaining
the result in Fig.~\ref{fig:snapshots_panda}. As in previous examples, all
methods improve $E_D$ and $E_K$ as $d$ increases. For $d>2$, the projection
method reduces the average kinematic error with respect to the basic, Baumgarte,
and PKT methods, and the subsequent application of the local coordinates method
eliminates such error up to numerical accuracy. Moreover, the total execution
time of first applying the projection method and then the local coordinates
methods is similar (or even shorter in some cases) than the execution time of
the alternative approaches. Thus, the proposed methods provide an efficient and
accurate pipeline for trajectory optimization in constrained robotic systems.

\section{Conclusions}  \label{sec:conclusion}

Existing trajectory optimization methods have difficulties when dealing with the
holonomic and nonholonomic constraints that appear in many problems of robotics.
As we have shown in this paper, these constraints confine the robot states to a
nonlinear manifold, but transcription errors easily produce trajectories that
drift away from this manifold. Such a drift typically translates into
unrealistic behaviour of the robot, and complicates, or even prevents, the
control of the trajectory a posteriori. To address these problems, this paper
has introduced two methods that cancel the drift without affecting, or even
reducing, the dynamic error of the trajectories. The two methods leverage
techniques from geometric integration on
manifolds~\cite{Hairer_NM2001,Hairer_SPRINGER2006,
	hairer1996solving,potra1991numerical,Potra_JSM1991}. The projection method is
simpler and typically faster, but it only cancels the drift at the knot points
of the trajectory. In contrast, the local coordinates method cancels the drift
all along the continuous-time trajectory. Both methods can employ polynomials of
arbitrary degree, which allows their application to hp-adaptive
meshing schemes~\cite{darby2011hp,Betts_SIAM2010}.

A key point in any trajectory optimizer is the quality of the initial guess of
the solution. To provide such a guess we have used the trajectories obtained
with the planner in~\cite{Bordalba_TRO2020}, which are not optimal in any
specific sense but satisfy all kinematic and dynamic constraints of the problem.
According to our experience, this guess is helpful to converge to the optimal
trajectories in reasonable times. For complex systems with many degrees of
freedom, however, a better strategy is to improve this guess with the projection
method (using a coarse mesh resolution or low-degree polynomials) and then use
it as a warm start for the local coordinates method. Thus, the two approaches
are complementary in this respect.

As usual in numerical optimal control, our final trajectories are local optima
in the set of trajectories homotopic to the initial guess. To obtain a
globally-optimal trajectory one should resort to asymptotically optimal planners
like those in~\cite{Li_IJRR2016,Hauser_TRO2016,Karaman_ICRA2011}. In this context, the
methods we propose could be used as a steering method for a global planner, so
the local connections could comply with both the kinematic and dynamic
constraints of the problem. This is a research direction that, we believe, is
worth exploring in the future.

\appendix

\section*{Computing a basis of the tangent space}
\label{app:tangent_basis}

To compute a basis of $\TX{\vr{x}_k}$, i.e., of the null space of
$\mt{F}_{\small \vr{x}}(\vr{x}_k)$, we can use the QR decomposition. Using this
decomposition, the $n_e \times n_x$ matrix $\mt{F}_{\small \vr{x}}(\vr{x}_k)$
can be expressed as
\begin{equation}
	\mt{F}_{\small \vr{x}}(\vr{x}_k)\tr = \mt{Q}_k \: \mt{R}_k, \label{eq:QRD}
\end{equation}
where $\mt{Q}_k$ is an $n_x \times n_x$ orthonormal matrix, so $\mt{Q}_k\tr
\mt{Q}_k = \mt{I}_{n_x}$, and $\mt{R}_k$ is an $n_x \times n_e$ upper triangular
matrix. From Eq.~\eqref{eq:QRD} we have that
\begin{equation}
	\mt{F}_{\small \vr{x}}(\vr{x}_k)\:\vr{Q}_k =\mt{R}_k\tr,
\end{equation}
which can be written in block form as
\begin{equation}
	\mt{F}_{\small \vr{x}}(\vr{x}_k)\:[\mt{V}_k \: \mt{U}_k]=[\mt{L}_k \: \mt{0}],
\end{equation}
where $\mt{V}_k$ includes the first $n_e$ columns of $\mt{Q}_k$, $\mt{U}_k$
includes the remaining $d_{\X}$ columns, and $\mt{L}_k$ is an $n_e \times n_e$
lower triangular matrix. Since $\mt{F}_{\small \vr{x}}(\vr{x}_k)\:
\mt{U}_k=\mt{0}$ and $\mt{U}_k\tr\mt{U}_k=\mt{I}_{d_{\X}}$, $\mt{U}_k$ provides
the desired orthonormal basis of $\TX{\vr{x}_k}$ .

Note that the basis $\mt{U}_k$ is not unique and typical implementations of the
QR decomposition apply column reordering with the aim of improving the numerical
stability of the procedure. Therefore, applying the process just described on
nearby points may produce significantly different bases, so the procedure does
not guarantee the continuity and smoothness of the outputs. This is inconvenient
for the optimization process, which requires the derivatives of~$\vr{U}_k$ with
respect to $\vr{x}_k$. Therefore, once~$\mt{U}_k$ is computed for a given point
$\vr{x}_k$, i.e., after the initialization of the optimization process, it is
more convenient to use Gram-Schmidt orthonormalization to update the basis
$\mt{U}_l$ for any point $\vr{x}_l$ close enough to the previous estimation
of~$\vr{x}_k$. In this process, the columns $\vr{u}_l^j$ of $\vr{U}_l$ for
$j=\{1,\ldots,d_{\X}\}$ are computed in sequence from the columns $\vr{u}_k^j$
of $\vr{U}_k$  in two steps. We first compute
\begin{equation*}
	\vr{w}_l^j=\vr{u}_k^j-\mt{A}_j\:\mt{A}_j\tr\:\vr{u}_k^j,
\end{equation*}
with $\mt{A}_j=[ \mt{F}_{\small \vr{x}}(\vr{x}_l)\tr \:\vr{u}_l^1 \ldots
\vr{u}_l^{j-1} ]$, and then obtain
\begin{equation*}
	\vr{u}_l^j=\frac{\vr{w}_l^j}{\| \vr{w}_l^j\|}.
\end{equation*}
The first step projects $\vr{u}_k^j$ to the null space of $\mt{A}_j$, i.e.,
of~$\mt{F}_{\small \vr{x}}(\vr{x}_l)$ and the vectors of $\mt{U}_l$ already
computed. The second step just normalizes the resulting vector. This process is
well-defined as long as none of the $\vr{w}_l^j$ vectors is zero, i.e., provided
$\mt{U}_k$ and $\mt{U}_l$ are relatively similar.

\bibliographystyle{IEEEtran}
\bibliography{IEEEabrv,references}

\end{document}